\documentclass[11pt]{article}

\newif\ifreview
\newif\ifarxiv
\newif\ifcamera

\newcommand{\arxivmode}{\reviewfalse\arxivtrue\camerafalse}

\arxivmode
\ifreview
  \usepackage[review]{acl}
\else
  \ifarxiv
    \usepackage[preprint]{acl}
  \else
    \usepackage[final]{acl}
  \fi
\fi

\usepackage{times}
\usepackage{latexsym}
\usepackage{microtype}
\usepackage{graphicx}
\usepackage{subcaption}
\usepackage{booktabs}
\usepackage{multirow}
\usepackage{xcolor}
\usepackage{colortbl}
\usepackage{cuted}
\usepackage{pifont}
\usepackage[most]{tcolorbox}

\usepackage{amsmath}
\usepackage{amssymb}
\usepackage{mathtools}
\usepackage{amsthm}
\usepackage{enumitem}
\usepackage{marvosym}
\usepackage{bbm}
\usepackage{listings}
\usepackage{booktabs}
\usepackage[normalem]{ulem}
\newcommand{\ul}[1]{\uline{#1}}

\usepackage{adjustbox}
\usepackage{booktabs}
\usepackage{xcolor}
\usepackage{xspace}

\newcommand{\nbf}[1]{{\noindent \textbf{#1}}}

\usepackage[capitalize,noabbrev]{cleveref}
\crefname{section}{Sec.}{Secs.}
\Crefname{section}{Appendix}{Appendices}
\crefname{table}{Table}{Tables}
\crefname{figure}{Fig.}{Figs.}
\crefname{algocf}{alg.}{algs.}
\Crefname{algocf}{Algorithm}{Algorithms}

\theoremstyle{plain}

\theoremstyle{definition}

\theoremstyle{remark}

\usepackage[T1]{fontenc}

\usepackage[utf8]{inputenc}

\usepackage{microtype}

\usepackage{inconsolata}

\usepackage[most]{tcolorbox}

\definecolor{promptframe}{HTML}{4F7C78}
\definecolor{promptbg}{HTML}{F6FAF9}
\definecolor{prompttitle}{HTML}{FFFFFF}
\definecolor{promptsep}{HTML}{CFE0DE}

\newtcolorbox{PromptBoxInner}[1]{%
  enhanced,
  colback=promptbg,
  colframe=promptframe,
  coltitle=prompttitle,
  fonttitle=\bfseries\footnotesize,
  title={#1},
  boxrule=0.6pt,
  arc=2pt,
  left=7pt, right=7pt, top=4pt, bottom=4pt,
  toptitle=3pt, bottomtitle=3pt,
  fontupper=\footnotesize,
  before upper={\setlength{\parskip}{0.4em}\setlength{\parindent}{0pt}},
}

\newenvironment{PromptBox}[3]{%
  \begin{figure*}[t]
  \centering
  \def\PromptBoxCaption{#2}%
  \def\PromptBoxLabel{#3}%
  \begin{minipage}{0.96\textwidth}
  \begin{PromptBoxInner}{#1}
}{%
  \end{PromptBoxInner}
  \end{minipage}
  \caption{\PromptBoxCaption}
  \label{\PromptBoxLabel}
  \end{figure*}
}

\title{WebRISE: Requirement-Induced State Evaluation for MLLM-Generated Web Artifacts}

\newcommand{\bench}{\texttt{WebRISE}\xspace}

\def\anonymousAuthorBlock{
First Author \\
Affiliation / Address line 1 \\
Affiliation / Address line 2 \\
Affiliation / Address line 3 \\
\texttt{email@domain} \\\And
Second Author \\
Affiliation / Address line 1 \\
Affiliation / Address line 2 \\
Affiliation / Address line 3 \\
\texttt{email@domain}
}

\def\realAuthorBlock{
    Yuxin Meng$^\textnormal{1,2}$\footnotemark[1] \qquad
    Yuhan Suo$^\textnormal{1,2}$\footnotemark[1] \qquad
    Junjie Wang$^\textnormal{1}$\footnotemark[1] \qquad
    Yuhan Sun$^\textnormal{3}$\footnotemark[1]
    \\
    \textbf{
    Yiyao Yu$^\textnormal{1}$ \qquad
    Ruixu Zhang$^\textnormal{1}$ \qquad
    Ruining Hu$^\textnormal{4}$ \qquad
    Yubin Wang$^\textnormal{2}$ \qquad
    Shouwei Ruan$^\textnormal{5}$
    }
    \\
    \textbf{
    Bin Wang$^\textnormal{2}$ \qquad
    Yuxiang Zhang$^\textnormal{2}$\footnotemark[2] \qquad
    Yujiu Yang$^\textnormal{1}$\footnotemark[2]
    }
    \vspace{-0.25em}
    \\\\
    \selectfont{$^\textnormal{1}$Tsinghua University} \quad
    \selectfont{$^\textnormal{2}$Huawei Noah's Ark Lab} \quad
    \selectfont{$^\textnormal{3}$East China Normal University}
    \\
    \selectfont{$^\textnormal{4}$Tongji University} \quad
    \selectfont{$^\textnormal{5}$Institute of Artificial Intelligence, Beihang University}
    \\
    \url{https://iigroup.github.io/WebRISE}
    \\
}

\ifreview
  \author{\anonymousAuthorBlock}
\else
  \author{\realAuthorBlock}
\fi

\begin{document}
\maketitle
\ifreview
\else
{
  \renewcommand{\thefootnote}{\fnsymbol{footnote}}
  \footnotetext[1]{Equal contribution.}
  \footnotetext[2]{Corresponding authors.}
  \footnotetext{Under Review.}
}
\fi
\begin{abstract}
Existing benchmarks for MLLM-generated web artifacts assess interaction through local evidence and miss the requirement-induced states and transitions that determine whether a page works.
We introduce \bench, which compiles task requirements into Interaction Contract Graphs (ICGs) of observable states, user-intent transitions, and DOM/visual assertions for implementation-agnostic browser execution.
\bench spans $442$ tasks across five input modalities (Text, Markdown, Sketch, Image, Video), with $5{,}495$ transitions and $5{,}271$ requirement checks that separate user-stated functions from implicit product-level constraints.
Across $14$ MLLMs, even the strongest model reaches only $65.6\%$ transition validity and $66.3\%$ requirement coverage, and visual quality is no proxy for behavior (Qwen3.6-35B-A3B on Markdown: $V{=}80.8$ yet $T{=}15.5$). 
Video gives the strongest interaction signal ($+10.6$\,pp implicit coverage over Text), while implicit constraints persist; defect injection shows ICG-based scoring detects state errors at $2$--$16\times$ the rate of checkpoint-style evaluation.
\end{abstract}

\section{Introduction}

Multimodal large language models (MLLMs) are increasingly asked to generate executable web artifacts from multimodal specifications, including textual requirements, Markdown structures, sketches, screenshots, and interaction videos~\citep{yin2024survey,si2025design2code,chen2025iwr,liu2026webcoderbench}.
This shift raises a basic benchmark question: \textit{when is a generated webpage usable, rather than merely visually plausible?}
In real use, a page can fail even when the expected controls are present: a filter may leave the item list unchanged, or a cart update may not propagate to the total price.
Evaluating MLLM-generated web artifacts therefore requires testing requirement-implied state transitions and state-consistency constraints, rather than only initial appearance or isolated action outputs.

\begin{figure}[!t]
\centering
\includegraphics[width=\linewidth]{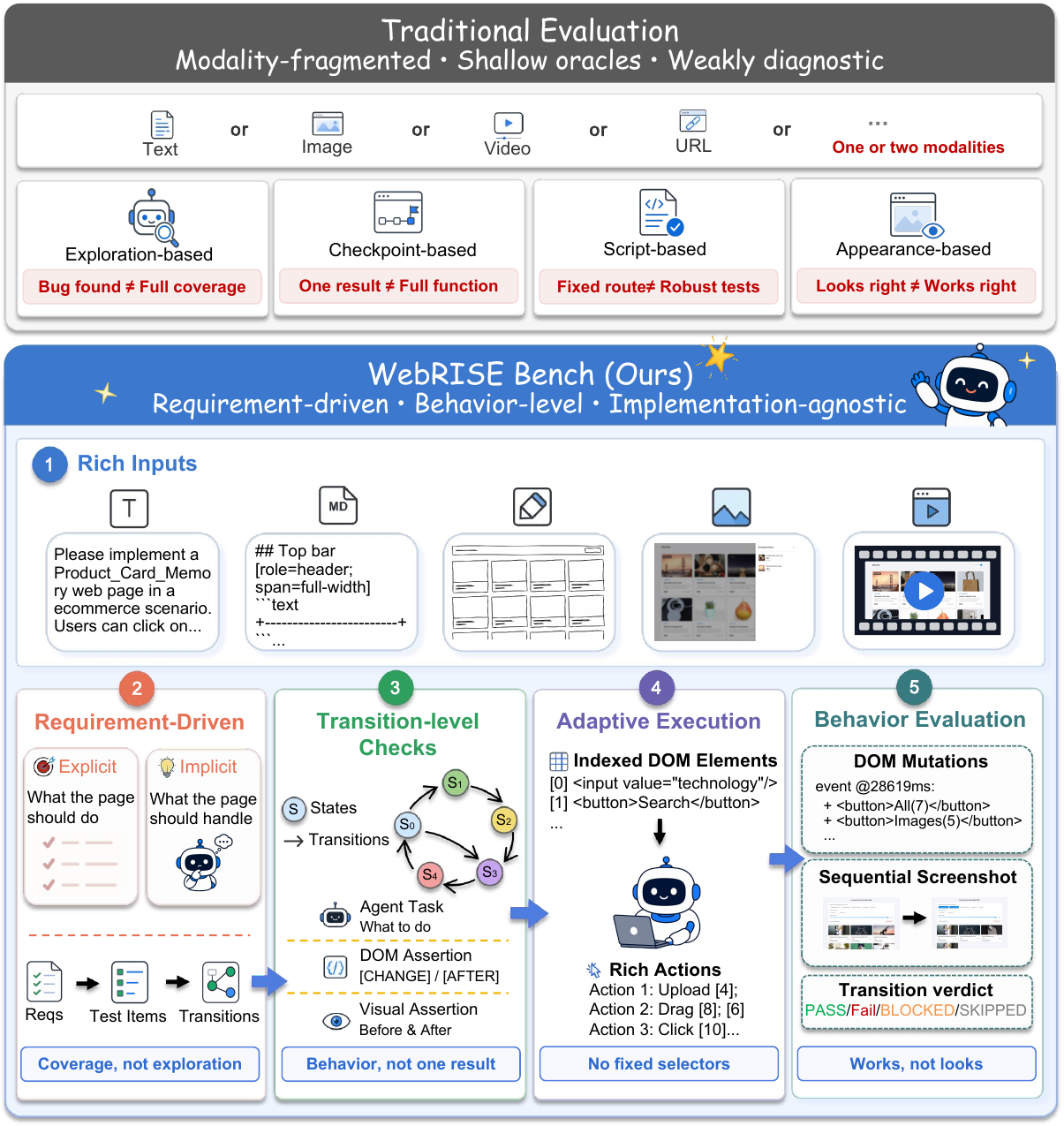}
\caption{
Overview of \bench.
Top: representative prior evaluation protocols often rely on modality-fragmented inputs and local evidence, such as appearance, scripts, checkpoints, or open-ended exploration.
Bottom: \bench evaluates generated web artifacts through a requirement-induced interaction contract: it supports five input modalities (\ding{182}), maps explicit and implicit requirements to test items and transitions (\ding{183}), defines DOM/visual transition checks (\ding{184}), executes them with a contract-guided agent (\ding{185}), and records transition-level verdicts with structured evidence (\ding{186}).
}
\label{fig:intro}
\vspace{-2em}
\end{figure}

Recent benchmarks for web, UI, and artifact generation have moved beyond static visual fidelity by incorporating interaction evidence, such as dynamic screenshots and MLLM-as-a-judge checklists~\citep{zhang2025artifactsbench}, predefined scripts~\citep{zhu2025frontendbench}, web-navigation agents~\citep{lu2026webgen}, real user requirements~\citep{liu2026webcoderbench}, and interaction videos~\citep{chen2025iwr}.
These efforts establish interaction as a central dimension of web generation evaluation.
However, existing protocols still tend to operationalize interaction through local evidence rather than requirement-level state obligations.
This creates two limitations.
(i) \textbf{Event-centric evaluation}: screenshots, script steps, video trajectories, or expected-result checkpoints can verify whether a local action produces a response, but they do not explicitly define which requirement-induced states and transitions should be covered.
(ii) \textbf{State-consistency gap}: a local response may pass even when the page violates cross-component, cross-view, or cross-step constraints, such as filter--pagination synchronization, count updates after deletion, or hidden-state preservation after navigation.
In short, existing benchmarks make interaction observable, but not yet fully enumerable or attributable as a requirement-induced state space.
\cref{fig:intro} summarizes this contrast.

To address these limitations, we introduce \bench, a benchmark that evaluates MLLM-generated web artifacts as \emph{requirement-induced observable state-transition conformance}.
\bench derives a finite interaction contract from task requirements, consisting of observable UI states, user-intent transitions, and DOM/visual assertions, and tests whether a generated page conforms to this contract under browser execution.
It is built on two design choices: \emph{requirement-conditioned state modeling}, which represents each task as an Interaction Contract Graph (ICG), and \emph{conformance-based diagnostic evaluation}, which links transition outcomes back to explicit requirements and implicit state-consistency constraints.

Concretely, \bench converts explicit and implicit requirements into Test Data Contracts and test items, then compiles them into an Interaction Contract Graph (ICG).
ICG states are requirement-relevant observable UI configurations rather than full DOM snapshots, while transient behaviors such as loading, saving, debounce, and temporary disabled states are verified as transition-level DOM evidence.
Each task is instantiated under Text, Markdown, Sketch, Image, and Video inputs, and models generate self-contained executable HTML pages.
During evaluation, the ICG specifies \emph{what} to verify, a contract-guided agent decides \emph{how} to execute each transition, and a DOM/visual dual oracle verifies process evidence and user-visible outcomes.
The resulting reports are aggregated into state-, transition-, and requirement-level diagnostics, including $S\%$, $T\%$, $Re\%$, $Ri\%$, and $R\%$.

\begin{table*}[!t]
\vspace{-10pt}
\small
\centering
\renewcommand{\arraystretch}{1.25}
\newcommand{\cmark}{\textcolor{teal}{\ding{51}}}
\newcommand{\xmark}{\textcolor{red!70!black}{\ding{55}}}
\resizebox{\textwidth}{!}{
\begin{tabular}{l l l ccc cc l}
\toprule
\textbf{Benchmark}
& \textbf{Verified Units}
& \textbf{Main Verdict}
& \textbf{Interact.}
& \textbf{Vision}
& \textbf{Safety}
& \textbf{Exp.Req}
& \textbf{Imp.Req}
& \textbf{Input Modality} \\
\midrule
WebCoderBench~\citep{liu2026webcoderbench}
& 1,572 reqs; 24 metrics
& Static evaluation
& \xmark & \cmark & \xmark
& \cmark & \xmark
& 2 (Text, Image) \\
VibeCodeBench~\citep{tran2026vibecodebench}
& 100 apps; 964 workflows
& Browser agent
& \cmark & \xmark & \xmark
& \cmark & \xmark
& 1 (Text) \\
Interaction2Code~\citep{xiao2025interaction2code}
& 127 pgs; 374 inter.
& Human evaluation
& \cmark & \cmark & \xmark
& \cmark & \xmark
& 1 (Image) \\
FrontendBench~\citep{zhu2025frontendbench}
& 148 tasks
& Script assertion
& \cmark & \xmark & \xmark
& \cmark & \xmark
& 1 (Text) \\
WebGen-Bench~\citep{lu2026webgen}
& 101 instr.; 647 cases
& Checkpoint
& \cmark & \cmark & \xmark
& \cmark & \xmark
& 1 (Text) \\
IWR-Bench~\citep{chen2025iwr}
& 113 tasks; 1,001 acts; 403 asserts
& MLLM assertion
& \cmark & \cmark & \xmark
& \cmark & \xmark
& 1 (Video) \\
\midrule
\textbf{\bench (ours)}
& \begin{tabular}[t]{@{}l@{}}\textbf{442 tasks; 5,271 reqs; 5,081 states;}\\\textbf{5,495 trans.; 12,441 asserts}\end{tabular}
& \textbf{DOM/VLM assertion}
& \cmark & \cmark & \cmark
& \cmark & \cmark
& \begin{tabular}[t]{@{}l@{}}\textbf{5} (Text, Markdown, Sketch,\\~~~Image, Video)\end{tabular} \\
\bottomrule
\end{tabular}
}
\caption{Comparison with related web generation benchmarks. Verdict: the mechanism used for pass/fail judgment. Exp.Req / Imp.Req: whether the benchmark includes explicit (user-stated) and implicit (unstated product-level) requirements separately. Input Modality: number of supported modalities with types listed.}
\label{tab:comparison}
\vspace{-10pt}
\end{table*}

We evaluate \bench on $442$ tasks, $5$ input modalities, and $14$ representative models, and obtain three main findings.
First, interactive web generation remains far from solved: even the strongest model, GPT-5.5, reaches only $T=65.6\%$ and $R=66.3\%$ under its best modality, leaving roughly one third of required transitions or requirement checks unsatisfied.
Second, multimodal specifications improve interaction quality, with Video being the strongest modality: compared with Text, it improves $T$, $R$, and $R_i$ by $8.8$, $8.3$, and $10.6$ percentage points, respectively.
Third, implicit state constraints remain a consistent bottleneck: explicit requirements are easier across models, and hard tasks are enriched with feedback, error, edge-state, and boundary-condition failures.
As an additional evaluator sanity check, defect injection on GT-validated pages shows that ICG-based evaluation detects $16/25$ injected state-related defects, compared with $8/25$ under a broad checkpoint-style WebGen criterion and $1/25$ under a strict one.

Our contributions are threefold:
\begin{itemize}[nosep,leftmargin=*]
\item We introduce \bench, a benchmark that reframes MLLM-generated web artifact evaluation as requirement-induced observable state-transition conformance, covering $442$ tasks, five input modalities, and explicit/implicit requirement contracts.
\item We develop a contract-guided evaluation protocol that represents each task with an Interaction Contract Graph, executes transitions with an adaptive browser agent, and verifies process and outcome evidence through DOM/visual oracles.
\item We conduct a large-scale evaluation of $14$ representative models, revealing that current systems remain far from solving interactive web generation, that Video provides the strongest interaction signal, and that implicit state constraints remain a major bottleneck.
\end{itemize}

\section{Related Work}

\nbf{MLLM-generated web artifacts.}

Multimodal large language models are increasingly moving from UI understanding and static code generation toward executable web artifact generation~\citep{yin2024survey}.
Early UI-to-code, design-to-code, and sketch-to-code studies mainly evaluate whether models can recover layout, visual structure, and front-end code from textual or visual specifications~\citep{si2025design2code,jain2019sketch2code,
periasami2026vision2code}.
Recent work further expands this setting to automated
functional testing~\citep{zhu2025frontendbench,lu2026webgen},
dynamic visual-interactive
evaluation~\citep{zhang2025artifactsbench},
real user requirements with interpretable
metrics~\citep{liu2026webcoderbench},
interactive webpage reconstruction from
video~\citep{chen2025iwr},
and agentic interactive
verification~\citep{xu2025webvia}.

This shift changes what should be evaluated.
For static pages or local components, visual fidelity, structural similarity, and code executability are natural targets.
For interactive web artifacts, however, the key question is whether the page responds correctly to user actions and preserves task-implied state constraints.
Accordingly, \bench evaluates MLLM-generated web artifacts as executable, stateful interfaces rather than merely rendered pages or code.

\nbf{Interactive web evaluation.}
Existing web generation benchmarks increasingly evaluate interaction through scripts, agents, visual judges, or demonstrated trajectories.
Script-based protocols such as FrontendBench~\citep{zhu2025frontendbench} provide reproducible functional checks but often depend on implementation-specific selectors or entry points.
Checkpoint-style protocols such as WebGen-Bench~\citep{lu2026webgen} use web-navigation agents to verify expected results, but still focus on local action--result pairs.
MLLM-judge and video-based protocols, such as ArtifactsBench~\citep{zhang2025artifactsbench} and IWR-Bench~\citep{chen2025iwr}, assess rendered evidence or trajectory reproduction.
Beyond generation benchmarks, agent-based web testing systems such as WebProber~\citep{ye2025ai} and UXAgent~\citep{lu2025uxagent} explore websites to identify bugs or usability issues.
These protocols make interaction observable, but typically operationalize it through scripts, checkpoints, trajectories, visual evidence, or exploration traces.
\bench instead formulates interaction evaluation as requirement conformance: an ICG defines requirement-linked states, transitions, and assertions, and an adaptive agent executes them on each generated page, supporting diagnosis beyond pass/fail outcomes.

\begin{figure*}
\centering
\includegraphics[width=\textwidth]{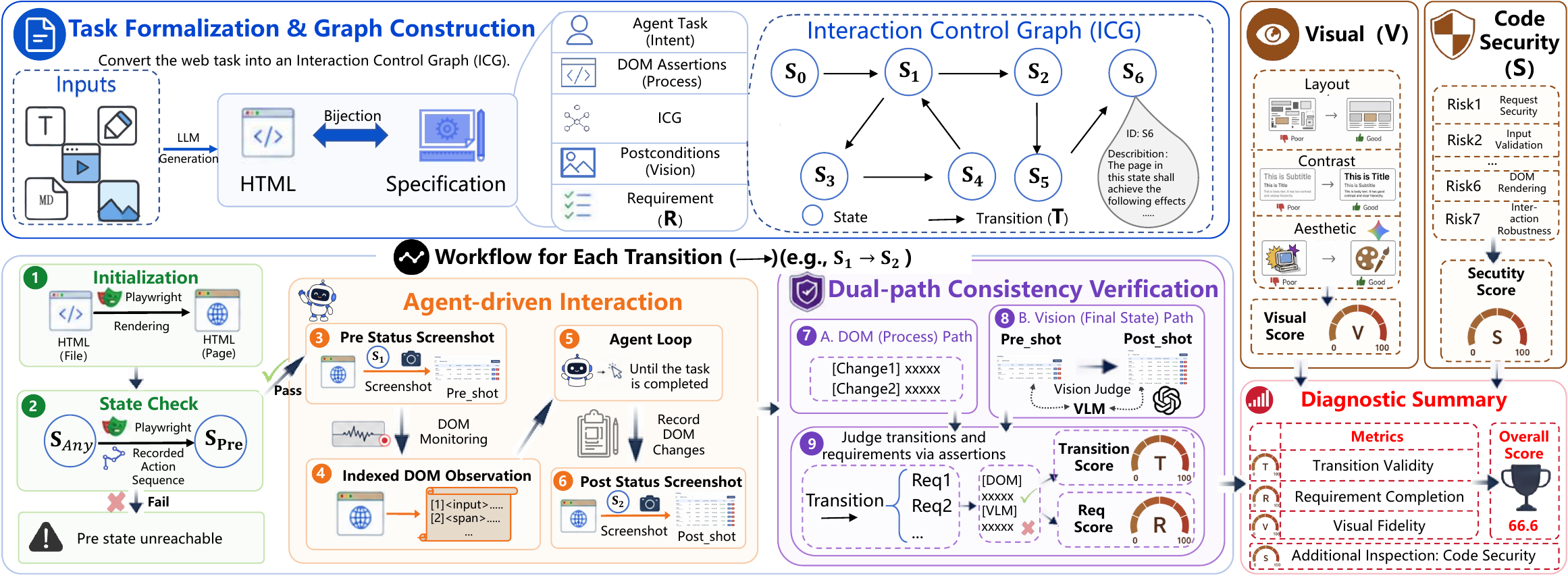}
\caption{
Overview of \bench.
\bench converts multimodal web generation tasks into Interaction Contract Graphs (ICGs), executes each state transition with a contract-guided agent, verifies process and outcome evidence with DOM/visual oracles, and aggregates transition-level verdicts into diagnostic scores.
}
\label{fig:overview}
\end{figure*}

\section{\bench: Benchmark Design}
\label{sec:benchmark_design}

\cref{fig:overview} summarizes the benchmark pipeline.
\bench converts task requirements into executable interaction contracts and evaluates generated HTML through browser-based conformance checks.

\subsection{Task Definition}
\label{sec:task_definition}

\bench evaluates whether an MLLM can generate an executable web artifact that satisfies the interaction behavior of a user-facing task.
For each task $\tau$, we define a requirement set $R_\tau$ and five modality-specific specifications $x_\tau^m$, where
{\small
\begin{equation}
m \in \mathcal{M}
=
\{\text{Text}, \text{Markdown}, \text{Sketch}, \text{Image}, \text{Video}\}.
\end{equation}
}
Given $x_\tau^m$, a model $f_\theta$ generates a self-contained HTML artifact:
{\small
\begin{equation}
    h_{\theta,\tau}^{m}=f_\theta(x_\tau^m).
\end{equation}
}
The artifact must be directly executable in a browser and include the required HTML, CSS, and JavaScript without external back-end services or manually prepared runtime state.

For each task, \bench derives a requirement-induced interaction contract $G_\tau$ from $R_\tau$.
The core evaluation asks whether $h_{\theta,\tau}^{m}$ satisfies $G_\tau$ under browser execution, rather than whether it matches a reference DOM, follows a fixed selector path, or reproduces a single visual snapshot.
Since $G_\tau$ is shared across modalities, \bench compares how textual, structural, visual, and temporal specifications affect generation of the same required interactive behavior.
Detailed modality construction procedures, prompt templates, and Image/Video specification rules are provided in~\cref{app:input_modalities}.

Ground-truth HTML pages validate contract executability and, when needed, provide Image/Video specifications, but are not treated as unique reference implementations.

\subsection{Requirement-Induced Interaction Contracts}
\label{sec:interaction_contracts}

For each task $\tau$, \bench derives an interaction contract from the requirement set $R_\tau$ and represents it as an Interaction Contract Graph (ICG):
{\small
\begin{equation}
    G_\tau = (S_\tau, T_\tau, \Phi_\tau, M_\tau).
\end{equation}
}

Here, $S_\tau$ denotes stable and replayable UI states, $T_\tau$ denotes user-intent-driven transitions, $\Phi_\tau$ denotes observable DOM/visual predicates, and $M_\tau$ maps requirements to test items, transitions, and assertions.

The states in $S_\tau$ are requirement-relevant observable UI configurations, rather than full DOM snapshots.
Transient effects such as loading indicators, saving states, toasts, debounce effects, and temporary disabled controls are not modeled as standalone states; they are attached to transitions as process-level predicates.
This keeps the state space finite and stable while preserving evidence for intermediate interaction behavior.

Each transition in $T_\tau$ specifies a user-intent state change, describing the desired outcome rather than a selector-level action sequence.
Predicates in $\Phi_\tau$ verify the transition through DOM evidence for structural or process-level signals and visual evidence for final user-visible outcomes, allowing the same contract to apply across diverse implementations.

The mapping $M_\tau$ connects transition-level evidence back to the original requirements.
Explicit requirements describe user-stated functional affordances, whereas implicit requirements capture product-level constraints such as state synchronization, boundary feedback, pagination reset, loading feedback, and stale-state removal.
Consequently, the contract specifies not only which interactions should be executed, but also how their evidence contributes to requirement-level evaluation.

\subsection{Contract Construction Pipeline}
\label{sec:construction_pipeline}

\bench constructs one interaction contract for each task and applies it to all model outputs across modalities.
The pipeline starts from expert-provided task materials and converts them into executable, requirement-attributable interaction contracts through four steps.

\nbf{Step 1: Expert-informed task collection.}
We design collection templates specifying the target domain, scenario, and expected web application setting.
Anonymous industry practitioners provide domain-grounded task materials, including user-facing requirements, representative interaction goals, and task-relevant data assumptions.
These materials serve as raw task sources, rather than executable evaluation specifications.

\nbf{Step 2: Requirement normalization.}
We normalize the collected materials into a requirement set $R_\tau$ for each task $\tau$.
Each set contains explicit requirements for user-stated functional affordances, such as search, filtering, sorting, dragging, and navigation, and implicit requirements for product-level interaction constraints, such as state synchronization, boundary feedback, pagination reset, loading feedback, and stale-state removal.

\nbf{Step 3: Test Data Contract and test items.}
From $R_\tau$, \bench derives a Test Data Contract specifying the minimal functional readiness for evaluation, such as initial data, filters, navigation entries, or loadable content, without constraining layout, DOM hierarchy, style, or exact element counts.
It derives test items that describe user-triggered behaviors and expected semantic outcomes, rather than CSS selectors, DOM paths, or click sequences.

\nbf{Step 4: ICG compilation.}
The Test Data Contract and test items are compiled into the Interaction Contract Graph $G_\tau$.
Stable configurations become states, user-triggered behaviors become transitions, and expected outcomes become DOM assertions or visual postconditions.
\bench also constructs the coverage mapping $M_\tau$, linking requirements to test items, transitions, and assertions.

This pipeline separates domain task authoring from executable evaluation design.
Practitioners provide realistic task content, while \bench converts it into an interaction contract that defines what should be evaluated; \cref{sec:evaluation_protocol} describes how the contract is executed on generated pages.

\subsection{Benchmark Statistics and Quality Control}
\label{sec:benchmark_statistics}

\begin{figure}[t]
\centering
\includegraphics[width=\linewidth]{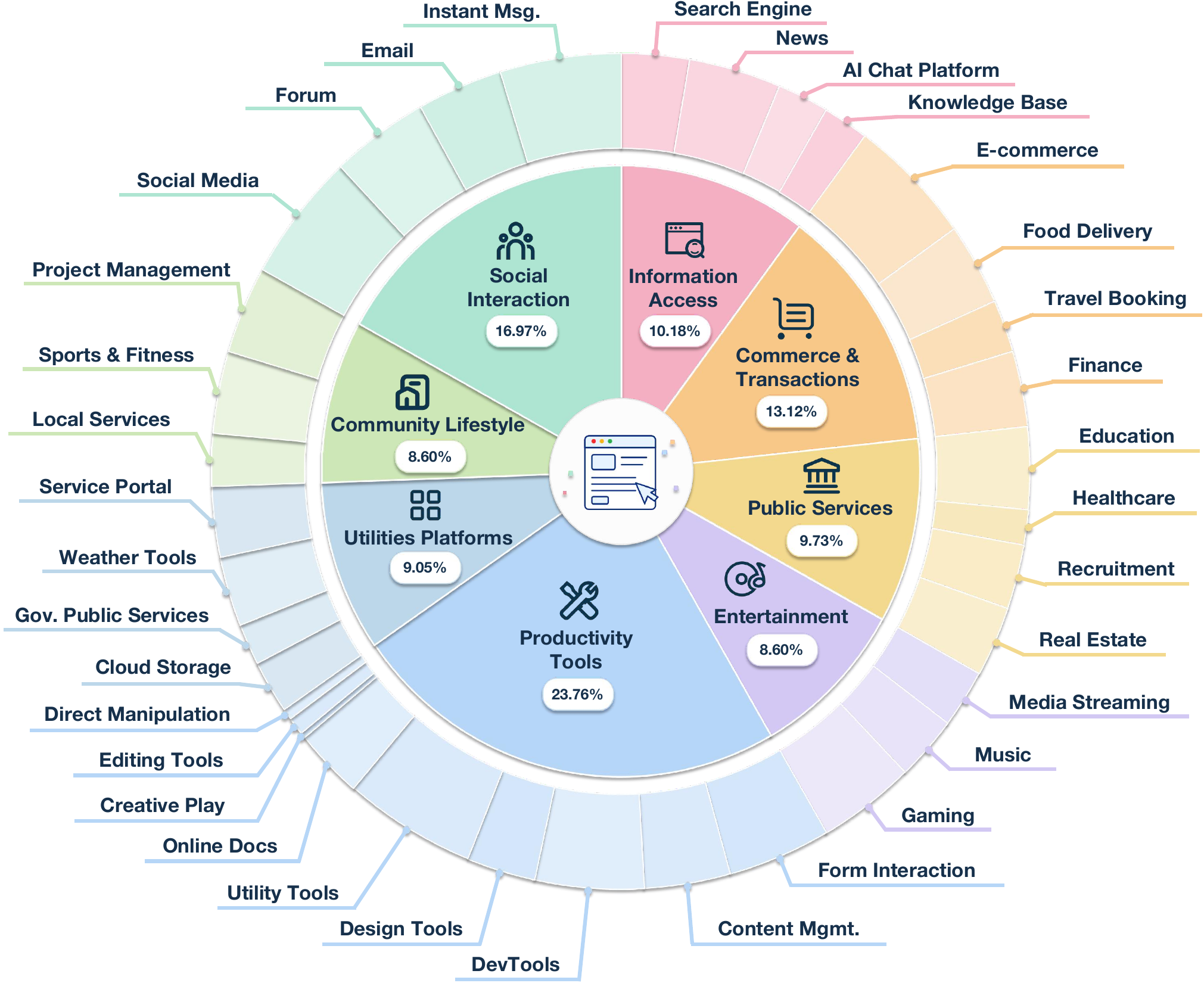}
\caption{
Domain and scenario distribution of \bench.
Tasks cover $8$ domains and $35$ scenarios, such as Productivity Tools ($23.76\%$) and Social Interaction ($16.97\%$).
}
\label{fig:domain-distribution}
\vspace{-18pt}
\end{figure}

\cref{fig:domain-distribution} shows that \bench spans diverse web application settings, with detailed construction statistics reported in~\cref{app:benchmark_statistics}.

After constructing each ICG, we validate it with a ground-truth HTML page generated from the full requirement set.
A task is retained only when the ground-truth page, the ICG, and the evaluator form a stable executable loop.
We also run schema checks over requirements, test items, states, transitions, assertions, and coverage mappings.
Human consistency validation is provided in~\cref{app:human_consistency}.

\section{Evaluation Protocol}
\label{sec:evaluation_protocol}

Given a generated HTML artifact $H$ and its Interaction Contract Graph $G_\tau$, \bench evaluates contract conformance under browser execution.
The ICG specifies \emph{what} to verify, while a contract-guided agent determines \emph{how} to execute each transition on the generated page.

\subsection{Protocol Overview}
\label{sec:protocol-overview}

Each transition is represented as
{\small
\begin{equation}
    t_j =
    (s_j^{\mathrm{from}}, s_j^{\mathrm{to}}, g_j, P_j,
    A^{\mathrm{dom}}_j, A^{\mathrm{vis}}_j),
\end{equation}
}
where $s_j^{\mathrm{from}}$ and $s_j^{\mathrm{to}}$ are source and target states, $g_j$ is the natural-language agent goal, $P_j$ is the precondition set, and $A^{\mathrm{dom}}_j$, $A^{\mathrm{vis}}_j$ are DOM assertions and visual postconditions.
This transition-level formulation supports branching state graphs and localizes evidence to requirement-linked state changes.

\cref{alg:eval-pipeline} summarizes the evaluation loop.
A transition is marked as \textsc{Pass} only if the source state is reachable, the agent completes the intended interaction, and all required DOM/visual checks hold.
The resulting reports are aggregated into the diagnostic metrics in \cref{sec:diagnostic_metrics}.

\begin{algorithm}[t]
\caption{Contract-Guided Evaluation}
\label{alg:eval-pipeline}
\small
\begin{algorithmic}[1]
\Require Page $H$, transitions $\mathcal{T}$, budget $K$, settle delay $\Delta$
\State Load $H$; initialize replay cache $\Pi \gets \emptyset$
\For{$t_j=(s_j^{\mathrm{from}},s_j^{\mathrm{to}},g_j,P_j,A^{\mathrm{dom}}_j,A^{\mathrm{vis}}_j)\in\mathcal{T}$}
    \State Restore $s_j^{\mathrm{from}}$ by replaying $\Pi$
    \If{restore fails}
        \State $o_j \gets \textsc{Skipped}$; record evidence; \textbf{continue}
    \EndIf
    \State Capture $\mathrm{img}_{\mathrm{pre}}$; check $P_j$
    \If{any precondition fails}
        \State $o_j \gets \textsc{Fail}$; record evidence; \textbf{continue}
    \EndIf
    \State Monitor DOM events; run agent on $g_j$ with budget $K$
    \State Wait $\Delta$; capture $\mathrm{img}_{\mathrm{post}}$; freeze event log $\mathcal{L}$
    \State $r_{\mathrm{dom}}\gets\Call{ScoreDOM}{A^{\mathrm{dom}}_j,\mathcal{L}}$
    \State $r_{\mathrm{vis}}\gets\Call{ScoreVisual}{A^{\mathrm{vis}}_j,\mathrm{img}_{\mathrm{pre}},\mathrm{img}_{\mathrm{post}}}$
    \State $o_j\gets\Call{Aggregate}{\mathrm{agent\ status},r_{\mathrm{dom}},r_{\mathrm{vis}}}$
    \If{$o_j=\textsc{Pass}$}
        \State Update $\Pi$ with the trajectory reaching $s_j^{\mathrm{to}}$
    \EndIf
    \State Record evidence $\mathcal{E}_j$
\EndFor
\end{algorithmic}
\end{algorithm}

\subsection{Contract-Guided Agent Execution}
\label{sec:agent_execution}

\bench uses an adaptive browser agent rather than a precompiled script.
At each step, the page is serialized into an indexed DOM observation containing interaction-relevant controls, state fields, newly appeared elements, scroll context, and editable text selections.
Because indices are regenerated after each action, execution depends on the current page state rather than fixed selectors or reference DOM paths.
For branching ICGs, source states are restored by replaying previously verified trajectories, which isolates transitions and separates unreachable states from executable contract violations.

\subsection{DOM/Visual Oracle and Evidence}
\label{sec:oracle_evidence}

Each transition is verified with a dual-channel oracle.
DOM assertions score process-level or element-level evidence from the event log, with \texttt{[CHANGE]} checking transient evidence during execution and \texttt{[AFTER]} checking the final stable DOM state.
Visual postconditions compare pre/post screenshots to verify final user-visible outcomes such as list updates, sorting changes, moved cards, opened panels, or empty states.
For auditability, \bench records the agent trace, DOM log, screenshots, assertion verdicts, and final transition outcome.
Details are provided in~\cref{app:evaluation_details}.

\begin{table*}[!t]
\scriptsize
\centering
\setlength{\tabcolsep}{2.2pt}
\renewcommand{\arraystretch}{1.12}
\resizebox{\textwidth}{!}{
\begin{tabular}{l|ccc|ccc|ccc|ccc|ccc|c}
\toprule
\multirow{2}{*}{\textbf{Model}}
& \multicolumn{3}{c|}{\textbf{Text}}
& \multicolumn{3}{c|}{\textbf{MD}}
& \multicolumn{3}{c|}{\textbf{Sketch}}
& \multicolumn{3}{c|}{\textbf{Image}}
& \multicolumn{3}{c|}{\textbf{Video}}
& \multirow{2}{*}{\textbf{Overall}} \\
\cmidrule(lr){2-4}\cmidrule(lr){5-7}\cmidrule(lr){8-10}\cmidrule(lr){11-13}\cmidrule(lr){14-16}
& $T$ & $R$ & $V$
& $T$ & $R$ & $V$
& $T$ & $R$ & $V$
& $T$ & $R$ & $V$
& $T$ & $R$ & $V$
& \\
\midrule
\multicolumn{17}{l}{\textit{Open-Source}} \\
\midrule
Qwen3.6-35B-A3B
& 26.8 & 30.5 & \underline{78.2}
& 15.5 & 19.2 & 80.8
& 41.2 & 45.4 & 77.0
& 46.6 & 49.6 & 71.7
& 49.5 & 52.2 & 72.8
& 50.5 \\
Qwen3.5-122B-A10B
& 38.0 & 41.2 & 56.8
& 42.5 & 45.9 & 72.0
& 38.0 & 42.3 & 74.0
& 40.2 & 43.8 & 70.7
& 42.8 & 47.1 & 71.3
& 51.1 \\
Qwen3.5-27B
& 36.3 & 40.0 & 59.9
& 41.7 & 45.5 & 72.1
& 38.6 & 42.7 & 76.8
& 42.6 & 46.7 & 70.6
& 43.1 & 46.9 & 71.8
& 51.7 \\
Qwen3.5-397B-A17B
& 45.7 & 49.2 & 64.8
& 51.1 & 54.5 & 75.7
& 46.8 & 50.5 & 78.9
& 48.4 & 51.4 & 72.8
& 49.3 & 52.8 & 72.1
& 57.6 \\
Kimi-K2.5
& \textbf{48.5} & \textbf{51.9} & 68.9
& \underline{57.0} & \underline{59.6} & 73.8
& \underline{47.8} & 50.4 & 79.9
& \underline{56.9} & \underline{59.1} & 72.6
& \underline{58.6} & \underline{60.3} & 72.9
& 61.2 \\
Qwen3.6-27B
& \underline{47.9} & \underline{50.9} & 75.3
& \textbf{57.5} & \textbf{60.1} & \underline{83.0}
& \textbf{50.4} & \textbf{53.3} & \textbf{87.2}
& 55.2 & 57.8 & \textbf{74.1}
& 54.2 & 57.2 & \textbf{74.1}
& \underline{62.5} \\
Kimi-K2.6
& 44.6 & 47.3 & \textbf{83.1}
& 51.7 & 54.9 & \textbf{87.1}
& \underline{47.8} & \underline{51.5} & \underline{86.3}
& \textbf{58.5} & \textbf{60.4} & \underline{73.2}
& \textbf{63.7} & \textbf{65.4} & \underline{73.5}
& \textbf{63.3} \\
\midrule
\multicolumn{17}{l}{\textit{Proprietary}} \\
\midrule
Claude Opus 4.6
& 43.3 & 45.5 & 56.6
& 54.3 & 56.3 & 73.9
& 52.3 & 55.0 & 72.2
& 57.7 & 59.5 & 70.2
& 52.6 & 54.9 & 70.7
& 58.3 \\
Gemini 3 Flash
& 44.7 & 48.2 & 71.9
& 50.0 & 54.1 & 79.3
& 46.1 & 49.3 & 85.4
& 54.1 & 57.5 & 72.4
& 45.6 & 48.5 & 70.8
& 58.5 \\
Claude Opus 4.7
& 48.8 & 50.9 & 68.3
& 54.5 & 56.5 & 76.2
& 49.7 & 52.4 & 77.4
& 57.0 & 58.5 & 70.5
& \underline{65.0} & \underline{66.1} & 72.7
& 61.6 \\
Gemini 3.1 Pro
& 50.7 & 53.6 & 69.7
& 58.9 & 61.5 & 79.2
& 52.2 & 54.9 & 84.8
& 54.5 & 57.1 & 72.2
& 52.0 & 54.9 & 71.6
& 61.9 \\
Qwen3.6-Plus
& 49.3 & 51.9 & 68.2
& 51.7 & 54.6 & 74.5
& 53.8 & 56.4 & \underline{86.3}
& 57.5 & 59.4 & \underline{73.8}
& 61.7 & 63.4 & \textbf{74.8}
& 62.5 \\
GPT-5.4
& \underline{59.7} & \underline{61.4} & \underline{78.4}
& \underline{60.5} & \underline{62.2} & \underline{79.8}
& \underline{57.8} & \underline{60.3} & \textbf{86.6}
& \underline{60.0} & \underline{62.1} & 71.5
& 63.1 & 64.8 & 73.7
& \underline{66.8} \\
GPT-5.5
& \textbf{60.3} & \textbf{62.3} & \textbf{85.6}
& \textbf{64.4} & \textbf{66.1} & \textbf{83.3}
& \textbf{60.6} & \textbf{62.9} & 86.1
& \textbf{61.8} & \textbf{63.4} & \textbf{74.1}
& \textbf{65.6} & \textbf{66.3} & \underline{73.9}
& \textbf{69.1} \\
\bottomrule
\end{tabular}
}
\caption{
Overall model performance on \bench across five input modalities.
We report transition validity ($T$), overall requirement coverage ($R$), and auxiliary visual quality ($V$); Overall is a compact average of $T$, $R$, and $V$ across modalities.
\textbf{Bold} and \underline{underline} denote the best and second-best results within each model group.
}
\label{tab:main-leaderboard}
\vspace{-10pt}
\end{table*}
 
\subsection{Diagnostic Metrics}
\label{sec:diagnostic_metrics}

\bench reports diagnostics as different projections of the same interaction contract.
After evaluation, each transition receives one outcome in
$\{\textsc{Pass}, \textsc{Fail}, \textsc{Blocked}, \textsc{Skipped}\}$.
Only \textsc{Pass} is counted as successful, which avoids giving credit to incomplete interactions or unreachable states.

\nbf{State and transition metrics.}
Let $S_\tau$ and $T_\tau$ denote the state and transition sets in $G_\tau$.
Let $S_\tau^{\mathrm{reach}}$ be the set of reached states, where the initial state is reachable only when its preconditions hold and any other state is reachable only through a passed incoming transition.
Let $T_\tau^{\mathrm{pass}}$ be the set of transitions marked as \textsc{Pass}.
We define:
{\small
\begin{equation}
    S\%(\tau)
    =
    \frac{|S_\tau^{\mathrm{reach}}|}{|S_\tau|}
    \times 100,
\end{equation}
}
{\small
\begin{equation}
    T\%(\tau)
    =
    \frac{|T_\tau^{\mathrm{pass}}|}{|T_\tau|}
    \times 100.
\end{equation}
}

Here, $S\%$ measures state reachability, while $T\%$ measures transition-level interaction correctness.

\nbf{Requirement coverage.}
Let $R_\tau^{\mathrm{exp}}$ and $R_\tau^{\mathrm{imp}}$ denote explicit and implicit requirements, with
$R_\tau = R_\tau^{\mathrm{exp}} \cup R_\tau^{\mathrm{imp}}$.
Using the coverage mapping $M_\tau$, each requirement $r$ is linked to the transitions and assertions that verify it.
We set $\mathrm{sat}(r)=1$ if all mapped checks for $r$ pass, and $0$ otherwise.
For any requirement subset
$\hat{R}\in\{R_\tau^{\mathrm{exp}},R_\tau^{\mathrm{imp}},R_\tau\}$,
we define:
{\small
\begin{equation}
    \mathcal{C}(\hat{R})
    =
    \frac{1}{|\hat{R}|}
    \sum_{r\in \hat{R}}
    \mathrm{sat}(r)
    \times 100.
\end{equation}
}
Applying $\mathcal{C}$ to $R_\tau^{\mathrm{exp}}$, $R_\tau^{\mathrm{imp}}$, and $R_\tau$ gives $Re\%$, $Ri\%$, and $R\%$, respectively.
$Re\%$ measures user-stated functional affordances, while $Ri\%$ measures implicit state-consistency constraints such as synchronization, boundary feedback, reset behavior, and stale-state removal.

\nbf{Aggregation.}
All metrics are computed at the task level and macro-averaged over tasks:
{\small
\begin{equation}
    \bar{q}(\theta,m)
    =
    \frac{1}{|\mathcal{D}|}
    \sum_{\tau\in\mathcal{D}}
    q(\theta,\tau,m),
\end{equation}
}
where $q \in \{S\%,T\%,Re\%,Ri\%,R\%\}$.
This prevents tasks with more transitions or assertions from dominating the aggregate score.

\section{Experiments and Findings}
\label{sec:experiments}

\subsection{Experimental Setup}
\label{sec:experimental_setup}

We evaluate \bench on $14$ representative models.
The model set includes $7$ open-weight models and $7$ proprietary models.
The open-weight models are Qwen3.5-27B~\citep{qwen35blog}, Qwen3.5-122B, Qwen3.5-397B, Qwen3.6-27B~\citep{qwen2026qwen36}, Qwen3.6-35B-A3B, Kimi K2.5~\citep{kimi25}, and Kimi K2.6~\citep{moonshot2026kimik26}.
The proprietary models are GPT-5.4~\citep{openai2026gpt54}, GPT-5.5~\citep{openai2026gpt55}, Claude Opus 4.6~\citep{anthropic2026claude46}, Claude Opus 4.7~\citep{anthropic2026claude47}, Gemini-3 Flash~\citep{google2025gemini3flash}, Gemini-3.1 Pro~\citep{google2026gemini31pro}, and Qwen3.6-Plus.

\subsection{Overall Model Performance}
\label{sec:overall_results}

\cref{tab:main-leaderboard} shows that interactive web artifact generation remains far from saturated.
Although GPT-5.5 achieves the highest compact Overall score, even its best modality, Video, reaches only $T=65.6$ and $R=66.3$, leaving roughly one third of required transitions or requirement checks unsatisfied.

\nbf{Proprietary models lead, but open-weight models remain competitive.}
GPT-5.5 and GPT-5.4 obtain the top two Overall scores, $69.1$ and $66.8$.
However, the gap is not determined solely by model access type.
Kimi-K2.6 achieves the best open-weight Overall score ($63.3$), surpassing several proprietary systems and performing especially well under Image and Video.
Qwen3.6-27B also reaches a competitive Overall score ($62.5$), with strong Markdown and Sketch results.
These trends suggest that modality handling and stateful interaction reasoning contribute substantially to model ranking.

\nbf{Visual quality is not a proxy for interaction correctness.}
High visual scores can coexist with weak executable behavior: Qwen3.6-35B-A3B obtains a strong Markdown visual score ($V=80.8$), but much lower interaction scores ($T=15.5$, $R=19.2$).
This mismatch reinforces the need to evaluate generated web artifacts through state transitions and requirement satisfaction, rather than visual plausibility alone.

\subsection{Analysis}

\begin{table}[t]
\centering
\small
\setlength{\tabcolsep}{4pt}
\renewcommand{\arraystretch}{1.1}
\begin{adjustbox}{max width=\linewidth}
\begin{tabular}{l c | l c}
\toprule
\textbf{Model} & \textbf{Pass (\%)} &
\textbf{Model} & \textbf{Pass (\%)} \\
\midrule
Gemini-3 Flash    & 24.9 & Qwen3.5-27B       & 29.8 \\
Kimi K2.5         & 28.0 & Qwen3.5-397B      & 30.0 \\
Qwen3.6-27B       & 28.0 & Qwen3.5-122B      & 30.2 \\
Kimi K2.6         & 28.2 & Gemini-3.1 Pro    & 31.0 \\
Qwen3.6-35B-A3B   & 28.3 & Claude Opus 4.7   & 31.6 \\
Claude Opus 4.6   & 28.8 & GPT-5.4           & \ul{34.9} \\
Qwen3.6-Plus      & 29.7 & GPT-5.5           & \textbf{41.3} \\
\bottomrule
\end{tabular}
\end{adjustbox}
\caption{
Auxiliary safety and robustness diagnostic results by model.
Pass rates are computed over applicable check instances; higher is better.
}
\label{tab:safety}
\vspace{-18pt}
\end{table}

\subsubsection{Safety and Robustness Diagnostics}
\label{sec:safety_diagnostics}

As an auxiliary diagnostic, we evaluate basic HTML safety and robustness checks.
\cref{tab:safety} shows uniformly low pass rates: even GPT-5.5 reaches only $41.3\%$, while most models cluster within $25$--$32\%$.
The flat model ranking and small cross-modality variation suggest that safer HTML generation is not automatically induced by stronger models or richer input specifications.

\begin{figure}[!t]
\centering
\includegraphics[width=\linewidth]{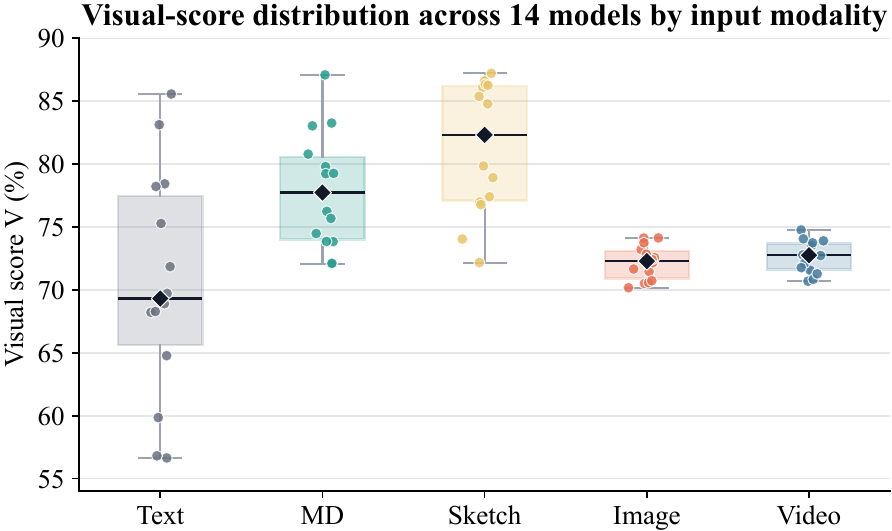}

\caption{
Visual-score distributions across input modalities.
Points denote models and boxes show distribution.
}
\label{fig:visual_score_distribution}
\vspace{-18pt}
\end{figure}

\subsubsection{Modality Effects}
\label{sec:modality_effects}

\cref{fig:visual_score_distribution} shows that visual quality and interaction performance follow different patterns.
Text has the largest cross-model variance, while Sketch obtains high visual scores due to strong spatial constraints from wireframes.
However, Image and Video have similar visual-score distributions, whereas Video leads in interaction-oriented metrics in~\cref{app:modality_results}.
This indicates that Video's advantage is better explained by temporal interaction evidence than by static visual fidelity, reinforcing that visual quality should remain an auxiliary signal.
The visual scoring procedure is described in~\cref{app:visual_evaluation}.

\begin{figure}[!t]
\centering
\includegraphics[width=\linewidth]{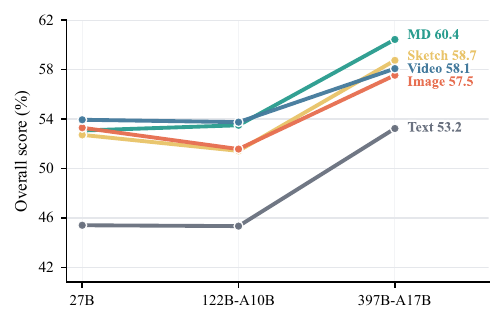}

\caption{
Scaling behavior of the Qwen3.5 family across input modalities.
Performance is largely flat from 27B to 122B-A10B, but increases sharply at 397B-A17B.
}
\label{fig:qwen35_scaling}
\vspace{-8pt}
\end{figure}

\subsubsection{Model Scaling Effects}

\cref{fig:qwen35_scaling} shows a non-linear scaling trend within the Qwen3.5 family: performance is largely flat from 27B to 122B-A10B, but improves clearly at 397B-A17B.
The gains are strongest under Text and Markdown, where layout, interaction logic, and state behavior must be inferred from weaker specifications.
This pattern suggests a scaling knee for stateful web artifact generation, where sufficient model capacity becomes important for jointly modeling layout, interaction logic, and state behavior.

\begin{table}[t]
\centering
\small
\setlength{\tabcolsep}{4pt}
\renewcommand{\arraystretch}{1.05}
\begin{tabular}{llcc}
\toprule
\textbf{Evaluator} & \textbf{Signal} & \textbf{Det.} & \textbf{DR(\%)} \\
\midrule
ICG       & $T < 100\%$    & 16/25 & 64.0 \\
WG-broad  & any non-YES    & 8/25  & 32.0 \\
WG-strict & $\geq$1 NO     & 1/25  & 4.0  \\
\bottomrule
\end{tabular}
\caption{
Defect injection meta-evaluation.
We compare ICG-based evaluation with checkpoint-style WebGen (WG) signals on 25 injected state-related defects.
Det. denotes detected defects and DR denotes detection rate.
ICG detects defects at $2\times$ the rate of WG under the broad criterion and $16\times$ under the strict criterion.
}
\label{tab:defect-injection}
\vspace{-18pt}
\end{table}

\subsubsection{Defect Injection Meta-Evaluation}
\label{sec:defect_injection}

To assess evaluator sensitivity, we inject state-related defects into GT-validated pages and rerun the same pipeline.
\cref{tab:defect-injection} shows that ICG-based evaluation detects substantially more defects than checkpoint-style WebGen signals, suggesting that explicit state-transition contracts are more sensitive to state corruptions missed by local checkpoints.
The remaining missed cases show that defect-sensitive evaluation is not yet exhaustive.

\subsubsection{Failure Attribution.}

\begin{figure}[!t]
\centering
\includegraphics[width=\linewidth]{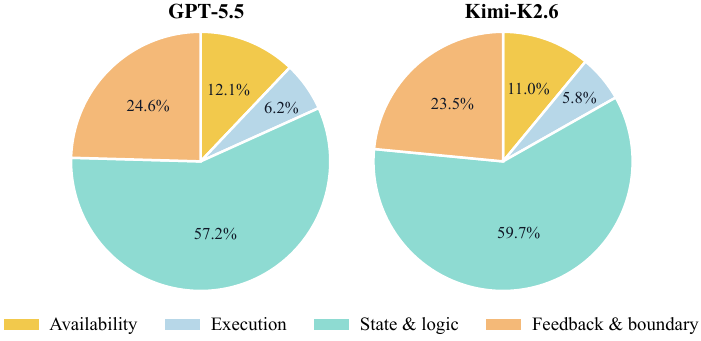}
\caption{Failure attribution (GPT-5.5 and Kimi-K2.6).}
\label{fig:fail_level1_gpt55_vs_kimik26}
\end{figure}

\cref{fig:fail_level1_gpt55_vs_kimik26} groups direct failed transitions into four functional error types.
GPT-5.5 and Kimi-K2.6 show similar profiles: \emph{State \& Logic} dominates, followed by \emph{Feedback \& Boundary}.
Therefore, many failures occur after required controls or interaction paths are exposed, indicating that the main bottleneck is maintaining correct state updates, result logic, validation behavior, and boundary feedback under user actions.

\begin{figure}
\centering
\includegraphics[width=\linewidth]{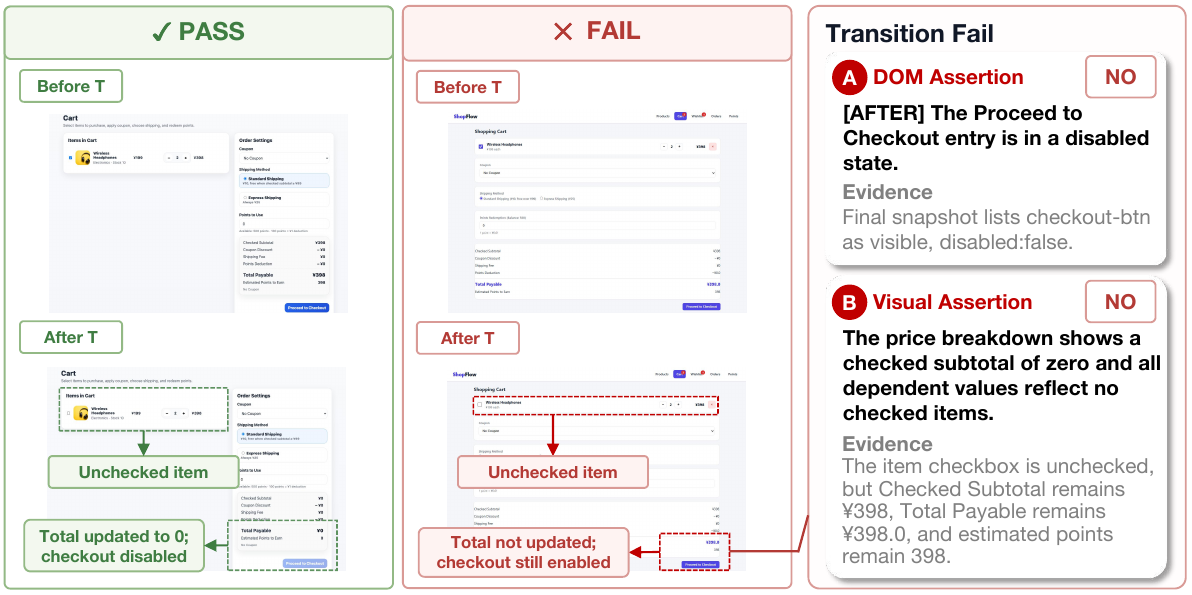}
\caption{
Case study of \bench's transition-level diagnosis on a shopping-cart interaction.
After the only checked item is unchecked, the passing artifact resets the totals to zero and disables checkout.
The failing artifact changes the item checkbox state but leaves the price breakdown and checkout availability stale; \bench localizes the error with failed DOM and visual assertions.
}
\label{fig:case-main}
\vspace{-18pt}
\end{figure}

\subsubsection{Case Study}
\cref{fig:case-main} illustrates transition-level diagnosis on a shopping-cart interaction.
The failing artifact accepts the user click but fails to propagate the resulting state change to dependent totals and checkout availability, exposing a state-consistency error rather than a click-execution failure.

\section{Conclusion}

We introduced \bench, a benchmark that evaluates MLLM-generated web artifacts through requirement-induced observable state-transition conformance.
\bench represents each task with an Interaction Contract Graph, enabling implementation-agnostic browser execution and state-, transition-, and requirement-level diagnostics over explicit functions and implicit state-consistency constraints.
Experiments on $442$ tasks, five input modalities, and $14$ models show that current systems remain far from solving interactive web generation: Video provides the strongest interaction signal, while implicit state constraints remain a persistent bottleneck.
These results highlight the need to evaluate generated web artifacts by requirement-level state behavior, rather than visual plausibility or isolated action success alone.

\section*{Limitations}

\bench focuses on self-contained HTML artifacts executed in a controlled browser environment.
This enables consistent comparison across models and modalities, but does not cover full production web systems involving back-end services, authentication, external APIs, persistent databases, multi-user concurrency, or long-lived sessions.
Accordingly, \bench should be interpreted as measuring front-end interaction conformance rather than deployment readiness.
A natural extension is to augment Interaction Contract Graphs with sandboxed API contracts, persistent data fixtures, and session-level state transitions.

\bench evaluates generated pages against requirement-induced interaction contracts.
Although the contracts are validated through ground-truth execution, schema checks, human consistency studies, and defect injection, their coverage is still bounded by the specified requirements, generated test items, and DOM/visual assertions.
Therefore, \bench provides diagnostic evidence of conformance to the defined interaction contract, rather than an exhaustive characterization of all possible user behaviors.
Future work can broaden coverage by expanding contract templates, adding richer defect suites, incorporating multiple evaluator agents and selectively auditing uncertain cases.

\section*{Ethical Considerations}

\bench is a diagnostic benchmark, not a deployable system. Contributors and annotators participated under informed consent with aggregated reporting. Because contributors are drawn primarily from a single region, regional product conventions shape what counts as expected interaction, and applications targeting other markets should treat our metrics as a baseline and extend the contract set with locale-specific affordances.
LLM-judge scoring is validated against human judgments ($\kappa = 0.74$, Appendix~\ref{app:human_consistency}) and defect injection, but remains susceptible to prompt sensitivity and API version drift; reported scores should be read as stable rank-orderings rather than absolute measurements. We release all judge prompts, configurations, and per-assertion verdicts to support independent re-scoring.

\bibliography{custom}

\clearpage
\appendix
\section*{Appendix}

\section{Additional Benchmark Details}
\label{app:benchmark_details}

\subsection{Benchmark Statistics}
\label{app:benchmark_statistics}

\cref{tab:app-eval-scale} reports additional construction statistics of \bench.
The benchmark contains $442$ tasks across $8$ domains and $35$ scenarios, instantiated under five input modalities.
At the interaction-contract level, it includes $5{,}081$ states, $5{,}495$ transitions, and $5{,}271$ requirement checks, covering both explicit user-stated requirements and implicit product-level constraints.

\subsection{Input Modality Construction}
\label{app:input_modalities}

\bench instantiates each task under five input modalities to simulate different specification conditions in practical web artifact generation.
The task $\tau$ and its interaction contract $G_\tau$ are fixed across modalities, while the input specification $x_\tau^m$ varies.
\cref{tab:app-modality-design} summarizes the information provided by each modality and its intended evaluation role.

\subsection{Human Consistency Validation}
\label{app:human_consistency}

We conduct human consistency validation to examine whether the constructed interaction contracts and automatic evaluators align with human judgements.
The validation covers two aspects: (i) the requirement-to-ICG construction and agent-based functional evaluation, and (ii) the modality-specific visual evaluation.
This study is used only as a consistency check for benchmark construction and evaluator reliability; it is not used to tune model outputs or change the main evaluation results.

\nbf{Annotation setup.}
We sample 300 interaction cases from \bench, stratified across domains, input modalities, and task difficulty levels.
Each interaction case contains the original task requirement, the corresponding test item or ICG transition, the generated page execution trace, and the automatic verdict.
Human annotators judge whether the transition correctly reflects the intended requirement and whether the generated page satisfies the expected functional interaction.
For the visual validation, we sample 300 generated HTML pages across the five input modalities.
Annotators evaluate visual quality according to the modality-specific criterion: single-page visual quality for Text, reference-page similarity for Image and Video, sketch similarity for Sketch, and Markdown-structure consistency for Markdown.

\nbf{Annotator disclosure and privacy.}
The annotators were informed about how the benchmark data were collected and how their annotations would be used in this research.
The annotation process does not require releasing private personal information.
For privacy reasons, we do not disclose additional identifying information about individual participants, such as names, employers, or detailed personal profiles.
All reported results are aggregated.

\nbf{Metrics.}
We report accuracy, mean absolute error (MAE), Spearman correlation, Pearson correlation, and Cohen's $\kappa$.
Accuracy and Cohen's $\kappa$ measure agreement on binary pass/fail judgements.
MAE and correlation metrics are computed over normalized scores when graded judgements are available.
For each validation setting, we compare the automatic result against the human-majority judgement, and also report human--human agreement as a reference.

\begin{table}[!t]
\centering
\small
\setlength{\tabcolsep}{5pt}
\begin{tabular}{@{}lr@{}}
\toprule
\textbf{Statistic} & \textbf{Value} \\
\midrule
Domains / Scenarios / Tasks & 8 / 35 / 442 \\
Input modalities & 5 \\
Task--modality instances & 2{,}210 \\
\midrule
States & 5{,}081 \\
Transitions & 5{,}495 \\
Requirement checks & 5{,}271 \\
\quad Explicit requirements & 2{,}276 \\
\quad Implicit requirements & 2{,}995 \\
\midrule
Avg.\ transitions / task & 12.4 \\
Avg.\ requirement checks / task & 11.9 \\
\bottomrule
\end{tabular}
\caption{
Benchmark construction statistics of \bench.
The table summarizes task coverage, modality instantiation, interaction-contract scale, and requirement-check composition.
}
\label{tab:app-eval-scale}
\end{table}

\begin{table*}[t]
\centering
\small
\setlength{\tabcolsep}{4pt}
\renewcommand{\arraystretch}{1.08}
\begin{tabular}{p{0.1\textwidth} p{0.4\textwidth} p{0.4\textwidth}}
\toprule
\textbf{Modality} & \textbf{Specification Signal} & \textbf{Evaluation Role} \\
\midrule
Text 
& Scenario description, task name, explicit requirements, and Test Data Contract. 
& Tests whether the model can implement user-stated functions and infer unstated product-level behavior from text-only requirements. \\
\addlinespace[2pt]

Markdown 
& Text specification augmented with symbolic layout structure, e.g., header, sidebar, main area, list, card, modal, or toolbar regions. 
& Tests whether structured layout cues help the model organize the page and ground interaction affordances. \\
\addlinespace[2pt]

Sketch 
& Low-fidelity wireframe preserving spatial layout while removing semantic content, copy, color, and visual details. 
& Tests whether the model can recover UI structure and affordances from coarse visual layout. \\
\addlinespace[2pt]

Image 
& Selected high-fidelity screenshot with visually inferable explicit requirements omitted from the accompanying text. 
& Tests whether visual grounding can recover visible components and complement text-only interaction logic. \\
\addlinespace[2pt]

Video 
& Interaction demonstration generated from a validated ICG transition chain.
& Tests whether the model can infer interaction flow, process feedback, and temporal state changes from demonstrations. \\
\bottomrule
\end{tabular}
\caption{
Input modalities in \bench.
All modalities share the same task and interaction contract, but expose different specification signals to the model.
}
\label{tab:app-modality-design}
\end{table*}

\begin{table*}[!t]
\centering
\small
\setlength{\tabcolsep}{5pt}
\renewcommand{\arraystretch}{1.06}
\begin{tabular}{lcccccc}
\toprule
\textbf{Comparison} & \textbf{\#Samples} & \textbf{Acc.$\uparrow$} & \textbf{MAE$\downarrow$} & \textbf{Spearman$\uparrow$} & \textbf{Pearson$\uparrow$} & \(\boldsymbol{\kappa}\)\textbf{$\uparrow$} \\
\midrule
Human (A) vs. Human (B) 
& 300 & 0.88 & 0.18 & 0.91 & 0.89 & 0.81 \\

Human majority vs. requirement-to-ICG construction
& 300 & 0.86 & 0.20 & 0.88 & 0.86 & 0.78 \\

Human majority vs. agent-based functional evaluation
& 300 & 0.84 & 0.23 & 0.86 & 0.84 & 0.74 \\
\bottomrule
\end{tabular}
\caption{
Human consistency validation for interaction-contract construction and functional evaluation.
The automatic requirement-to-ICG construction and agent-based evaluation are compared with human-majority judgements, with human--human agreement reported as a reference.
}
\label{tab:human_consistency_interaction}
\end{table*}

\nbf{Interaction consistency.}
As shown in \cref{tab:human_consistency_interaction}, the automatic requirement-to-ICG construction achieves 0.86 accuracy and a Cohen's $\kappa$ of 0.78 against the human majority.
The agent-based functional evaluator achieves 0.84 accuracy, 0.86 Spearman correlation, 0.84 Pearson correlation, and a Cohen's $\kappa$ of 0.74.
These scores are close to the human--human agreement, suggesting that both the constructed interaction contracts and the automatic functional evaluation provide stable signals for requirement-level interaction correctness.

\begin{table*}[!t]
\centering
\small
\setlength{\tabcolsep}{5pt}
\renewcommand{\arraystretch}{1.06}
\begin{tabular}{lcccccc}
\toprule
\textbf{Comparison / Setting} & \textbf{\#Samples} & \textbf{Acc.$\uparrow$} & \textbf{MAE$\downarrow$} & \textbf{Spearman$\uparrow$} & \textbf{Pearson$\uparrow$} & \(\boldsymbol{\kappa}\)\textbf{$\uparrow$} \\
\midrule
Human (A) vs. Human (B) 
& 300 & 0.85 & 0.17 & 0.88 & 0.86 & 0.77 \\

Human majority vs. Text visual evaluator
& 60 & 0.82 & 0.24 & 0.83 & 0.80 & 0.70 \\

Human majority vs. Image/Video visual evaluator
& 120 & 0.81 & 0.25 & 0.81 & 0.79 & 0.69 \\

Human majority vs. Sketch/Markdown visual evaluator
& 120 & 0.80 & 0.26 & 0.79 & 0.77 & 0.67 \\

Human majority vs. overall visual evaluator
& 300 & 0.81 & 0.25 & 0.80 & 0.78 & 0.69 \\
\bottomrule
\end{tabular}
\caption{
Human consistency validation for modality-specific visual evaluation.
The visual evaluator is compared with human-majority judgements under modality-specific criteria, including single-page visual quality for Text, reference-page similarity for Image/Video, sketch similarity for Sketch, and structure consistency for Markdown.
}
\label{tab:human_consistency_visual}
\end{table*}

\nbf{Visual consistency.}
\cref{tab:human_consistency_visual} reports the consistency of the visual evaluator.
The overall visual evaluator obtains 0.81 accuracy, 0.80 Spearman correlation, 0.78 Pearson correlation, and a Cohen's $\kappa$ of 0.69 against the human majority.
The agreement is slightly lower than the functional interaction evaluation, which is expected because visual assessment involves more subjective judgement.
Nevertheless, the results indicate that the visual evaluator provides a stable auxiliary signal for modality-specific layout quality, visual consistency, and reference alignment.

\section{Additional Evaluation Protocol Details}
\label{app:evaluation_details}

\subsection{Agent Observation and Action Space}
\label{app:agent_interface}

\bench provides the evaluation agent with a compact, action-oriented view of the live webpage rather than the full HTML document.
At each interaction step, the browser state is converted into an indexed DOM observation that exposes only interaction-relevant elements and state fields.
Each actionable element receives an ephemeral \texttt{index}, which is local to the current observation and regenerated after the next browser action.
This design allows the agent to act on the current page state without relying on persistent CSS selectors, fixed DOM paths, or reference-specific implementation details.

\nbf{Indexed DOM observation.}
For each serialized element, \bench records its tag, accessibility role, visible text, key attributes, and interaction states.
Typical fields include \texttt{placeholder}, \texttt{value}, \texttt{href}, \texttt{type}, \texttt{checked}, \texttt{selected}, \texttt{expanded}, \texttt{pressed}, \texttt{disabled}, \texttt{aria-disabled}, and \texttt{pointer-events}.
For structured or stateful widgets, the observation additionally records option lists, slider values, scroll offsets, and cursor or selection ranges for editable regions.
These fields support fine-grained interactions such as selecting text spans, operating custom dropdowns, restoring scroll context, and performing drag-and-drop transitions.

\nbf{Non-standard components.}
Generated pages often implement interactive elements with custom DOM structures rather than native controls.
Therefore, \bench includes not only native buttons, links, inputs, selects, and text areas, but also elements with interactive ARIA roles, non-negative \texttt{tabindex}, event listeners, pointer or text cursors, \texttt{contenteditable}, or hover-revealed subtrees.
Newly appeared elements are marked through cross-step DOM diffing, and hidden or non-interactable elements are explicitly annotated.
This makes the agent interface robust to diverse MLLM-generated implementations while avoiding the cost and brittleness of exposing the full DOM.

\nbf{Action space.}
The agent action space covers common web operations and interaction-heavy behaviors.
It includes pointer actions, keyboard and text actions, form-control actions, spatial actions, and navigation/lifecycle actions:
\texttt{Click}, \texttt{Hover}, \texttt{Type}, \texttt{Clear}, \texttt{PressKey}, \texttt{SelectOption}, \texttt{ToggleCheck}, \texttt{SetSliderValue}, \texttt{Scroll}, \texttt{DragAndDrop}, \texttt{UploadFile}, \texttt{CanvasClickAt}, \texttt{Back}, \texttt{Refresh}, \texttt{WaitFor}, and \texttt{Done}.
We additionally support \texttt{SelectText} for selecting contiguous text spans inside \texttt{input}, \texttt{textarea}, and \texttt{contenteditable} regions.
These actions allow \bench to evaluate interactions that cannot be expressed by simple click/type scripts, including anchored text editing, drag-and-drop reordering, file upload, slider control, canvas selection, and browser navigation recovery.

\subsection{DOM and Visual Assertion Scoring}
\label{app:assertion_scoring}

\bench scores each transition with two complementary assertion channels.
DOM assertions operate on structured browser evidence, including the initial DOM snapshot, the final DOM snapshot, and the event log collected during agent execution.
Visual postconditions operate on pre- and post-interaction screenshots.
This separation lets \bench capture transient process evidence and element-level states through DOM signals, while using visual evidence for final user-visible outcomes.

\nbf{DOM assertion scoring.}
Each DOM assertion is prefixed with a temporal operator.
\texttt{[CHANGE]} requires the condition to hold at some point during the execution timeline, and is used for transient signals such as loading, saving, progress, debounce, confirmation feedback, or temporary disabled states.
\texttt{[AFTER]} requires the condition to hold in the final stable DOM state, and is used for persistent outcomes such as selected filters, disabled controls, removed items, restored buttons, or updated ARIA states.

To reduce free-form interpretation, \bench applies deterministic priority rules for common state predicates.
For non-interactivity, the scorer first checks \texttt{pointer-events: none}, then native \texttt{disabled} or \texttt{aria-disabled="true"}, and then state-indicative class tokens such as \texttt{disabled}, \texttt{inactive}, \texttt{locked}, or \texttt{readonly}.
For selection or activation, the scorer prioritizes \texttt{aria-selected}, \texttt{aria-pressed}, and \texttt{aria-checked}, followed by class tokens such as \texttt{selected}, \texttt{active}, \texttt{highlighted}, or \texttt{current}.
For expansion, it uses \texttt{aria-expanded} and visibility changes in the corresponding container subtree.

Element localization uses visible text, role, \texttt{aria-label}, placeholder, attributes, and child-structure summaries.
When multiple candidates match the target and the evidence is insufficient to disambiguate them, the scorer returns \textsc{Uncertain} rather than selecting a target arbitrarily.
Only \textsc{Yes} is treated as passing when aggregating assertion-, transition-, and requirement-level scores.

\nbf{Visual postcondition scoring.}
Visual postconditions compare the screenshots before and after a transition.
They are written as behavioral conditions rather than pixel-level constraints, so different implementations can pass if they satisfy the same user-visible semantics.
Typical postconditions include list updates, sorting changes, panel expansion, drag-and-drop placement, empty-state display, stale-state removal, and visible value updates.

The visual scorer uses before/after differences to judge the requested semantic change.
For conditional assertions, it first determines which branch applies from the screenshots and evaluates only that branch.
If relevant content is clipped by the viewport or a scrollable container, the scorer relies only on fully visible evidence.
For search or filter assertions, an empty result may pass when the filter is visibly active and the page shows a valid empty state.
Ambiguous or unsupported visual evidence is marked as \textsc{Uncertain}, and does not count as a passing assertion.

\subsection{Transition Outcomes and Evidence}
\label{app:transition_evidence}

Each evaluated transition receives one of four outcomes.
\textsc{Pass} indicates that the source state is reachable, the agent completes the intended interaction, and all required DOM/visual checks pass.
\textsc{Fail} indicates that the transition is executable but at least one required assertion or postcondition is violated.
\textsc{Blocked} indicates that the agent cannot complete the interaction within the budget, typically because the required affordance is absent, hidden, or non-functional.
\textsc{Skipped} indicates that the source state cannot be restored, usually because a prerequisite transition failed or the replay path is unavailable.
This taxonomy separates contract violations from execution failures and prevents a single upstream defect from being counted repeatedly across downstream transitions.

For auditability, \bench stores a structured evidence bundle for every transition.
The bundle includes the transition identifier, source and target state descriptors, the natural-language agent goal, pre- and post-interaction screenshots, the agent action trace, the DOM event log, initial and final DOM snapshots, per-assertion verdicts, the final transition outcome, and the replay path when state replay is used.
Each per-assertion record stores the verdict, supporting evidence fragments, and scorer version.
The evidence bundle allows each reported error to be traced to the relevant phase, such as source-state restoration, agent execution, DOM assertion scoring, visual postcondition scoring, or replay.
It also supports manual auditing, phase-level error analysis, and defect-injection meta-evaluation.

\subsection{Additional Metric Details}
\label{app:metric_details}

In addition to the main metrics, \bench records test-item-level and assertion-level signals for diagnostic analysis.
These signals are not used as primary leaderboard metrics, but help localize errors between user-facing behaviors, transition checks, and individual evidence channels.

\nbf{Test-item coverage.}
A test item corresponds to a user-triggered behavior and its expected semantic outcome.
Using the coverage mapping $M_\tau$, each test item is linked to the transitions and assertions that verify it.
We mark a test item as satisfied only when all mapped transitions and required assertions pass:
{\small
\begin{equation}
    TI\%(\tau)
    =
    \frac{1}{|I_\tau|}
    \sum_{i\in I_\tau}
    \mathrm{sat}(i)
    \times 100,
\end{equation}
}
where $I_\tau$ is the set of test items for task $\tau$ and $\mathrm{sat}(i)\in\{0,1\}$.
Because test items are closer to user-facing behaviors than raw transitions, $TI\%$ is mainly used for qualitative error analysis.

\nbf{Assertion-level verdicts.}
Each DOM assertion and visual postcondition receives a verdict in
$\{\textsc{Yes}, \textsc{No}, \textsc{Uncertain}\}$.
Only \textsc{Yes} is treated as passing when aggregating assertion-, transition-, test-item-, and requirement-level scores.
\textsc{No} indicates contradicted evidence, while \textsc{Uncertain} indicates insufficient or ambiguous evidence.
This conservative rule prevents ambiguous observations from inflating final scores.

\nbf{Aggregation convention.}
Unless otherwise specified, model- and modality-level scores are computed by macro-averaging task-level scores.
This gives each task equal weight and prevents tasks with more transitions, assertions, or requirements from dominating aggregate results.
Assertion-level and test-item-level metrics are used for debugging, case studies, and failure attribution, while the main paper focuses on state reachability, transition validity, and explicit/implicit requirement coverage.

\nbf{Compact overall score.}
For leaderboard readability, we report an auxiliary Overall score:
{\small
\begin{equation}
    O(\theta)
    =
    \frac{1}{|\mathcal{M}|}
    \sum_{m\in\mathcal{M}}
    \frac{T(\theta,m)+R(\theta,m)+V(\theta,m)}{3},
\end{equation}
}
where $V$ is the modality-specific auxiliary visual score.
Overall is used only as a compact summary; the primary analysis relies on the diagnostic interaction and requirement metrics, especially $T\%$, $R_e\%$, $R_i\%$, and $R\%$.

\subsection{Visual Quality Evaluation Details}
\label{app:visual_evaluation}

\bench reports visual quality as an auxiliary signal, complementary to executable interaction metrics.
The visual evaluator combines three components: layout structure, color accessibility, and perceptual aesthetics, with modality-specific aggregation.

\nbf{Layout and color.}
The layout module performs coarse-grained block modeling over the rendered page, measuring alignment, structural clarity, and floating-element artifacts.
When a visual reference is available, it also measures cross-page structural consistency using row-level signatures and grid-distribution similarity.
The color module checks text contrast against WCAG thresholds ($\geq$4.5:1 for normal text and $\geq$3:1 for large text), and for Image/Video additionally compares palette and contrast-profile similarity to the reference page.

\nbf{Aesthetics.}
A VLM-based scorer evaluates screenshots along high-level perceptual dimensions, including whitespace balance, recurring-element consistency, hierarchy clarity, and overall polish.
This complements rule-based layout and color checks with visual judgments that are difficult to encode deterministically.

\nbf{Modality-specific aggregation.}
For Text, aesthetics is the primary signal, with layout and color used as auxiliary checks.
For Markdown and Sketch, structural similarity to the reference specification receives the largest weight, supplemented by aesthetics.
For Image and Video, layout fidelity and color reproduction relative to the reference page are primary, with aesthetics as a secondary signal.
All visual scores are macro-averaged across tasks, and full model-by-modality visual scores are reported in~\cref{tab:app-visual-model-modality}.

\begin{table}[t]
\centering
\scriptsize
\setlength{\tabcolsep}{2.4pt}
\renewcommand{\arraystretch}{1.05}
\resizebox{\linewidth}{!}{
\begin{tabular}{lcccccc}
\toprule
\textbf{Model} & \textbf{Avg.} & \textbf{Text} & \textbf{MD} & \textbf{Sketch} & \textbf{Image} & \textbf{Video} \\
\midrule
Kimi-K2.6 & 80.6 & 83.1 & 87.1 & 86.3 & 73.2 & 73.5 \\
GPT-5.5 & 80.6 & 85.6 & 83.3 & 86.1 & 74.1 & 73.9 \\
Qwen3.6-27B & 78.7 & 75.3 & 83.0 & 87.2 & 74.1 & 74.1 \\
GPT-5.4 & 78.0 & 78.4 & 79.8 & 86.6 & 71.5 & 73.7 \\
Qwen3.6-35B-A3B & 76.1 & 78.2 & 80.8 & 77.0 & 71.7 & 72.8 \\
Gemini 3 Flash & 76.0 & 71.9 & 79.3 & 85.4 & 72.4 & 70.8 \\
Qwen3.6-Plus & 75.5 & 68.2 & 74.5 & 86.3 & 73.8 & 74.8 \\
Gemini 3.1 Pro & 75.5 & 69.7 & 79.2 & 84.8 & 72.2 & 71.6 \\
Kimi-K2.5 & 73.6 & 68.9 & 73.8 & 79.9 & 72.6 & 72.9 \\
Claude Opus 4.7 & 73.0 & 68.3 & 76.2 & 77.4 & 70.5 & 72.7 \\
Qwen3.5-397B-A17B & 72.9 & 64.8 & 75.7 & 78.9 & 72.8 & 72.1 \\
Qwen3.5-27B & 70.2 & 59.9 & 72.1 & 76.8 & 70.6 & 71.8 \\
Qwen3.5-122B-A10B & 69.0 & 56.8 & 72.0 & 74.0 & 70.7 & 71.3 \\
Claude Opus 4.6 & 68.7 & 56.6 & 73.9 & 72.2 & 70.2 & 70.7 \\
\bottomrule
\end{tabular}
}
\caption{
Auxiliary visual-quality scores by model and input modality.
Scores are reported on a 0--100 scale and macro-averaged across tasks.
}
\label{tab:app-visual-model-modality}
\end{table}

\begin{table}[t]
\centering
\small
\setlength{\tabcolsep}{5pt}
\renewcommand{\arraystretch}{1.06}
\begin{tabular}{lccc}
\toprule
\textbf{Judge model} & \textbf{\#Pairs} & \textbf{Detected} \\
\midrule
GPT-5.4    & 100 & 97 / 100  \\
GPT-5-mini & 100 & 95 / 100  \\
\bottomrule
\end{tabular}
\caption{
Judge-model robustness validation on 100 sampled GT/defect-injected HTML pairs.
}
\label{tab:judge-robustness}
\end{table}

\section{Additional Experimental Details and Results}
\label{app:additional_results}

\subsection{Evaluation Judge Configuration}
\label{app:eval_judge_models}

We use GPT-5-mini for transition-level DOM assertion and visual postcondition scoring, and Gemini-3-Flash-Preview for auxiliary visual-quality scoring.
The same judge configuration is applied to all evaluated models, tasks, and modalities.

To verify that the lighter transition-level judge does not reduce defect sensitivity, we compare GPT-5-mini with GPT-5.4 on 100 sampled GT/defect-injected HTML pairs.
Each pair contains a GT-validated page that passes the ICG-based evaluation and a corresponding defect-injected variant that introduces a controlled interaction fault.
As shown in \cref{tab:judge-robustness}, GPT-5-mini remains close to GPT-5.4 on this sampled control set.

\subsection{Additional Modality Analysis}
\label{app:modality_results}

\begin{table}[!t]
\centering
\small
\setlength{\tabcolsep}{3.2pt}
\renewcommand{\arraystretch}{1.05}
\begin{tabular}{lccccccc}
\toprule
\multirow{2}{*}{\textbf{Mod.}} 
& \multirow{2}{*}{$T$} 
& \multicolumn{3}{c}{\textbf{Req. Coverage}} 
& \multirow{2}{*}{$R$} 
& \multirow{2}{*}{$V$} 
& \multirow{2}{*}{\textbf{Overall}} \\
\cmidrule(lr){3-5}
& & $R_e$ & $R_i$ & $\Delta{\downarrow}$ & & & \\
\midrule
Text   & 46.0 & 56.4 & 43.0 & 13.5 & 48.9 & 70.4 & 55.1 \\
MD     & 50.8 & 61.1 & 47.6 & 13.5 & 53.7 & \underline{77.9} & \underline{60.8} \\
Sketch & 48.8 & 60.1 & 45.4 & 14.7 & 52.0 & \textbf{81.4} & 60.7 \\
Image  & \underline{53.6} & \textbf{62.8} & \underline{50.8} & \underline{12.0} & \underline{56.2} & 72.2 & 60.7 \\
Video  & \textbf{54.8} & \underline{61.3} & \textbf{53.6} & \textbf{7.7} & \textbf{57.2} & 72.6 & \textbf{61.5} \\
\bottomrule
\end{tabular}
\caption{
Average modality-level performance on \bench across all evaluated models and tasks.
$T$ denotes transition validity; $R_e$ and $R_i$ denote explicit and implicit requirement coverage; $\Delta=R_e-R_i$ is the explicit--implicit gap; $R$ denotes overall requirement coverage; $V$ is the auxiliary visual score; and Overall is the mean of $T$, $R$, and $V$.
Bold and underlined values indicate the best and second-best results in each column.
}
\label{tab:app-modality-results}
\end{table}

\cref{tab:app-modality-results} reports modality-level averages across all evaluated models and tasks.
Video achieves the strongest interaction-oriented performance, leading in transition validity ($T$), implicit requirement coverage ($R_i$), and overall requirement coverage ($R$), while reducing the explicit--implicit gap to $7.7$ points.
This suggests that temporal demonstrations are especially helpful for recovering state changes and implicit product-level behavior.
Image obtains the highest explicit requirement coverage ($R_e=62.8$) and closely follows Video on $T$ and $R$, indicating that high-fidelity visual grounding helps models recover visible components and initial interface state.
By contrast, Sketch obtains the highest auxiliary visual score ($V=81.4$), but lags behind Image and Video on interaction and requirement metrics.
This indicates that visual organization alone is not a reliable proxy for executable interaction correctness.

\begin{table*}[t]
\scriptsize
\centering
\setlength{\tabcolsep}{1.25pt}
\renewcommand{\arraystretch}{1.08}
\resizebox{\textwidth}{!}{
\begin{tabular}{l|cccccc|cccccc|cccccc|cccccc|cccccc|c}
\toprule
\multirow{2}{*}{\textbf{Model}}
& \multicolumn{6}{c|}{\textbf{Text}}
& \multicolumn{6}{c|}{\textbf{MD}}
& \multicolumn{6}{c|}{\textbf{Sketch}}
& \multicolumn{6}{c|}{\textbf{Image}}
& \multicolumn{6}{c|}{\textbf{Video}}
& \multirow{2}{*}{\textbf{Overall}} \\
\cmidrule(lr){2-7}\cmidrule(lr){8-13}\cmidrule(lr){14-19}\cmidrule(lr){20-25}\cmidrule(lr){26-31}
& $S$ & $T$ & $R_e$ & $R_i$ & $R$ & $V$
& $S$ & $T$ & $R_e$ & $R_i$ & $R$ & $V$
& $S$ & $T$ & $R_e$ & $R_i$ & $R$ & $V$
& $S$ & $T$ & $R_e$ & $R_i$ & $R$ & $V$
& $S$ & $T$ & $R_e$ & $R_i$ & $R$ & $V$
& \\
\midrule
\multicolumn{32}{l}{\textit{Open-Source}} \\
\midrule
Qwen3.6-35B-A3B
& 31.8 & 26.8 & 36.6 & 25.9 & 30.5 & \underline{78.2}
& 20.0 & 15.5 & 22.3 & 16.7 & 19.2 & 80.8
& 47.1 & 41.2 & 54.5 & 38.1 & 45.4 & 77.0
& 51.8 & 46.6 & 56.7 & 43.8 & 49.6 & 71.7
& 53.3 & 49.5 & 57.7 & 47.9 & 52.2 & 72.8
& 50.5 \\
Qwen3.5-122B-A10B
& 42.8 & 38.0 & 48.9 & 35.2 & 41.2 & 56.8
& 47.5 & 42.5 & 54.1 & 39.3 & 45.9 & 72.0
& 43.4 & 38.0 & 49.7 & 36.2 & 42.3 & 74.0
& 45.4 & 40.2 & 50.9 & 38.1 & 43.8 & 70.7
& 47.0 & 42.8 & 51.7 & 43.5 & 47.1 & 71.3
& 51.1 \\
Qwen3.5-27B
& 41.4 & 36.3 & 47.3 & 34.3 & 40.0 & 59.9
& 46.9 & 41.7 & 53.5 & 38.8 & 45.5 & 72.1
& 44.6 & 38.6 & 50.7 & 36.5 & 42.7 & 76.8
& 47.7 & 42.6 & 53.6 & 41.2 & 46.7 & 70.6
& 47.2 & 43.1 & 51.0 & 43.4 & 46.9 & 71.8
& 51.7 \\
Qwen3.5-397B-A17B
& 51.2 & 45.7 & 57.2 & 42.8 & 49.2 & 64.8
& 56.2 & 51.1 & 62.3 & 48.2 & 54.5 & 75.7
& 52.5 & 46.8 & \underline{60.1} & 42.8 & 50.5 & 78.9
& 53.2 & 48.4 & 57.7 & 46.3 & 51.4 & 72.8
& 53.3 & 49.3 & 56.8 & 49.4 & 52.8 & 72.1
& 57.6 \\
Qwen3.6-27B
& \underline{52.7} & \underline{47.9} & \underline{58.4} & \underline{44.8} & \underline{50.9} & 75.3
& \textbf{62.2} & \textbf{57.5} & \textbf{67.3} & \textbf{54.3} & \textbf{60.1} & \underline{83.0}
& \textbf{55.6} & \textbf{50.4} & \textbf{60.9} & \textbf{47.2} & \textbf{53.3} & \textbf{87.2}
& 60.3 & 55.2 & 64.8 & 52.0 & 57.8 & \textbf{74.1}
& 58.5 & 54.2 & 61.4 & 53.4 & 57.2 & \textbf{74.1}
& \underline{62.5} \\
Kimi-K2.5
& \textbf{53.5} & \textbf{48.5} & \textbf{59.4} & \textbf{46.1} & \textbf{51.9} & 68.9
& \underline{61.9} & \underline{57.0} & \textbf{67.3} & \underline{53.5} & \underline{59.6} & 73.8
& 52.8 & \underline{47.8} & 58.3 & 44.0 & 50.4 & 79.9
& \underline{61.3} & \underline{56.9} & \underline{65.2} & \underline{54.0} & \underline{59.1} & 72.6
& \underline{62.2} & \underline{58.6} & \underline{65.0} & \underline{56.5} & \underline{60.3} & 72.9
& 61.2 \\
Kimi-K2.6
& 49.4 & 44.6 & 54.2 & 41.8 & 47.3 & \textbf{83.1}
& 56.5 & 51.7 & \underline{62.9} & 48.4 & 54.9 & \textbf{87.1}
& \underline{53.0} & \underline{47.8} & 58.8 & \underline{45.6} & \underline{51.5} & \underline{86.3}
& \textbf{63.2} & \textbf{58.5} & \textbf{66.6} & \textbf{55.4} & \textbf{60.4} & \underline{73.2}
& \textbf{67.1} & \textbf{63.7} & \textbf{68.4} & \textbf{62.6} & \textbf{65.4} & \underline{73.5}
& \textbf{63.3} \\
\midrule
\multicolumn{32}{l}{\textit{Proprietary}} \\
\midrule
Gemini 3 Flash
& 49.7 & 44.7 & 56.2 & 41.5 & 48.2 & 71.9
& 55.3 & 50.0 & 63.2 & 47.0 & 54.1 & 79.3
& 51.2 & 46.1 & 57.7 & 42.7 & 49.3 & 85.4
& 59.5 & 54.1 & 64.7 & 51.5 & 57.5 & 72.4
& 49.9 & 45.6 & 53.5 & 44.2 & 48.5 & 70.8
& 58.5 \\
Claude Opus 4.6
& 47.9 & 43.3 & 53.1 & 39.5 & 45.5 & 56.6
& 58.8 & 54.3 & 63.3 & 50.6 & 56.3 & 73.9
& 57.5 & 52.3 & 63.6 & 48.0 & 55.0 & 72.2
& 62.1 & 57.7 & 65.9 & 54.2 & 59.5 & 70.2
& 55.7 & 52.6 & 58.4 & 51.7 & 54.9 & 70.7
& 58.3 \\
Gemini 3.1 Pro
& 55.6 & 50.7 & 61.1 & 47.5 & 53.6 & 69.7
& 63.6 & 58.9 & 69.5 & 54.9 & 61.5 & 79.2
& 56.8 & 52.2 & 62.5 & 48.8 & 54.9 & 84.8
& 59.1 & 54.5 & 63.3 & 51.9 & 57.1 & 72.2
& 55.8 & 52.0 & 58.9 & 51.5 & 54.9 & 71.6
& 61.9 \\
Qwen3.6-Plus
& 54.2 & 49.3 & 58.6 & 46.6 & 51.9 & 68.2
& 56.7 & 51.7 & 62.6 & 48.0 & 54.6 & 74.5
& 58.8 & 53.8 & 63.8 & 50.6 & 56.4 & \underline{86.3}
& 61.7 & 57.5 & 66.0 & 54.0 & 59.4 & \underline{73.8}
& 65.1 & 61.7 & 68.3 & 58.9 & 63.4 & \textbf{74.8}
& 62.5 \\
Claude Opus 4.7
& 53.4 & 48.8 & 57.6 & 45.8 & 50.9 & 68.3
& 58.6 & 54.5 & 63.1 & 51.2 & 56.5 & 76.2
& 54.3 & 49.7 & 59.3 & 46.9 & 52.4 & 77.4
& 61.3 & 57.0 & 64.5 & 53.9 & 58.5 & 70.5
& \underline{67.9} & \underline{65.0} & \textbf{70.0} & \underline{62.8} & \underline{66.1} & 72.7
& 61.6 \\
GPT-5.4
& \underline{64.6} & \underline{59.7} & \underline{70.3} & \underline{54.3} & \underline{61.4} & \underline{78.4}
& \underline{65.2} & \underline{60.5} & \underline{70.5} & \underline{55.4} & \underline{62.2} & \underline{79.8}
& \underline{62.7} & \underline{57.8} & \underline{70.2} & \underline{52.4} & \underline{60.3} & \textbf{86.6}
& \underline{64.5} & \underline{60.0} & \underline{68.7} & \underline{56.6} & \underline{62.1} & 71.5
& 66.1 & 63.1 & 68.4 & 61.6 & 64.8 & 73.7
& \underline{66.8} \\
GPT-5.5
& \textbf{65.1} & \textbf{60.3} & \textbf{71.1} & \textbf{55.3} & \textbf{62.3} & \textbf{85.6}
& \textbf{69.1} & \textbf{64.4} & \textbf{73.6} & \textbf{59.8} & \textbf{66.1} & \textbf{83.3}
& \textbf{65.3} & \textbf{60.6} & \textbf{71.6} & \textbf{56.0} & \textbf{62.9} & 86.1
& \textbf{66.4} & \textbf{61.8} & \textbf{69.8} & \textbf{58.0} & \textbf{63.4} & \textbf{74.1}
& \textbf{68.4} & \textbf{65.6} & \underline{69.4} & \textbf{63.5} & \textbf{66.3} & \underline{73.9}
& \textbf{69.1} \\
\bottomrule
\end{tabular}
}
\caption{Full model $\times$ modality results with state reachability ($S$), transition validity ($T$), explicit ($R_e$) and implicit ($R_i$) requirement coverage breakdown, and modality-specific visual scores.}
\label{tab:app-full-model-modality}
\end{table*}

\begin{table}[t]
\centering
\small
\setlength{\tabcolsep}{3.5pt}
\renewcommand{\arraystretch}{1.05}
\begin{tabular}{l|ccc|ccc}
\toprule
\multirow{2}{*}{\textbf{Mod.}}
& \multicolumn{3}{c|}{\textbf{Hard50}}
& \multicolumn{3}{c}{\textbf{Easy50}} \\
\cmidrule(lr){2-4}\cmidrule(lr){5-7}
& $T$ & $R_i$ & $R$
& $T$ & $R_i$ & $R$ \\
\midrule
Text   & 24.3 & 22.1 & 27.0 & 68.7 & 64.6 & 71.4 \\
MD     & 25.5 & 23.7 & 28.4 & 73.7 & 70.4 & 77.1 \\
Sketch & 22.4 & 21.0 & 25.7 & 73.0 & 69.0 & 75.9 \\
Image  & \underline{31.6} & \underline{31.1} & \underline{34.7} & \underline{74.5} & \underline{71.2} & \underline{77.3} \\
Video  & \textbf{36.5} & \textbf{37.2} & \textbf{39.5} & \textbf{76.1} & \textbf{72.9} & \textbf{78.7} \\
\bottomrule
\end{tabular}
\caption{
Performance on the R-based Hard50 and Easy50 splits by input modality.
Hard50 and Easy50 are selected as the 50 tasks with the lowest and highest model-averaged overall requirement coverage ($R$), respectively.
Video leads on both splits, with a larger advantage on Hard50, especially for implicit requirement coverage ($R_i$).
}
\label{tab:difficulty}
\end{table}

\subsection{Difficulty and Failure Attribution}
\label{sec:difficulty_analysis}
\nbf{Failure-type taxonomy.}
To analyze where functional failures occur along the interaction implementation chain, we group direct failed transitions into four functional error types.
\textbf{Availability} captures whether the page provides the required entry point, control, or interaction flow for completing the task.
\textbf{Execution} captures whether a user action takes effect when the relevant control or input area is present.
\textbf{State \& Logic} captures whether the page correctly updates state, data rules, target content, visual status, and context after an action.
\textbf{Feedback \& Boundary} captures whether the page correctly handles validation, disabled states, loading, errors, confirmations, and empty states.

To understand whether low scores arise from uniformly harder tasks or from qualitatively different failure modes, we analyze the R-based Hard50 and Easy50 splits from both performance and failure-attribution perspectives.

\cref{tab:difficulty} compares the R-based Hard50 and Easy50 splits by input modality.
The performance gap is large across all modalities, confirming that Hard50 captures genuinely difficult interaction tasks rather than small metric fluctuations.
Video remains the strongest modality on both splits, but its margin is much larger on Hard50: compared with Image, Video improves $T$, $R_i$, and $R$ by $4.9$, $6.1$, and $4.8$ points on Hard50, but only by $1.6$, $1.7$, and $1.4$ points on Easy50.
This suggests that dynamic interaction evidence is especially useful when tasks require non-trivial state transitions and implicit behavior recovery.

\begin{figure}[!t]
\centering
\includegraphics[width=\linewidth]{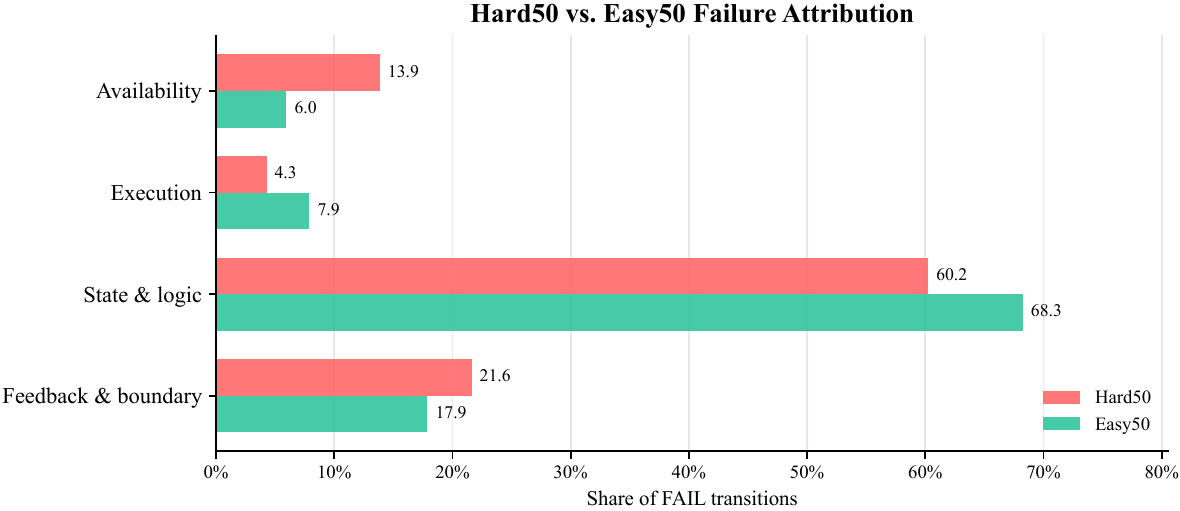}

\caption{
Failure-family attribution on the R-based Hard50 and Easy50 splits.
State and logic errors dominate both splits, while Hard50 shows larger shares of availability and feedback/boundary failures.
}
\label{fig:hard_easy_failure_attribution_level1}
\end{figure}

\cref{fig:hard_easy_failure_attribution_level1} further shows that the two splits expose different failure profiles.
State and logic errors dominate both Hard50 and Easy50, indicating that stateful result logic remains the central bottleneck.
However, Hard50 contains higher shares of availability failures and feedback/boundary failures, suggesting that difficult tasks often fail before or around the interaction boundary: required affordances may be missing, states may be unreachable, or edge-state feedback may be incomplete.
By contrast, Easy50 failures are more concentrated in state and logic errors, meaning that models often expose a basic interaction path but still fail to maintain the correct result logic or state consistency.

\subsection{Full Model \texorpdfstring{$\times$}{x} Modality Results}

\cref{tab:app-full-model-modality} reports the full model-by-modality results.

\begin{table*}[t]
\centering
\small
\setlength{\tabcolsep}{4pt}
\renewcommand{\arraystretch}{1.08}
\begin{tabular}{p{0.22\textwidth} c p{0.32\textwidth} p{0.32\textwidth}}
\toprule
\textbf{Error Pattern} & \textbf{N} & \textbf{Why WebGen Misses It} & \textbf{Example} \\
\midrule
Record loss under repeated operations
& 4
& WebGen verifies the latest successful action, but does not check whether earlier records remain visible.
& Multiple prize-wheel spins appear successful, but only the latest prize is kept in the winning records. \\
\addlinespace[2pt]

Cross-feature side effects
& 5
& WebGen evaluates the target feature locally, without checking whether unrelated UI state or components are changed.
& Editing one saved link succeeds, but other saved links have their URLs silently corrupted. \\
\addlinespace[2pt]

Draft state loss across navigation
& 2
& WebGen checks that pages and forms are reachable, but not whether intermediate user inputs persist across navigation.
& A leave type is selected, then resets after the user visits another module and returns. \\
\addlinespace[2pt]

Missing explicit-action gating
& 1
& WebGen checks that an output can be produced, but not whether it waits for the intended user trigger.
& A calculator result updates immediately after editing the input, before the user clicks execute. \\
\addlinespace[2pt]

Pre/post state inconsistency
& 1
& WebGen accepts confirmation feedback or a results page, without exact before--after value comparison.
& A poll vote is confirmed, but the selected option's count does not increase. \\
\bottomrule
\end{tabular}
\caption{
ICG-only error patterns among cases where WebGen marks all test items as \textsc{Yes}.
Counts are computed over 13 defect-injected cases detected only by ICG.
}
\label{tab:icg-only-error-patterns}
\end{table*}

\begin{table*}[t]
\small
\centering
\setlength{\tabcolsep}{3.5pt}
\renewcommand{\arraystretch}{1.05}
\begin{tabular}{llcc}
\toprule
\textbf{Check ID} & \textbf{Interpretation} & \textbf{Checks} & \textbf{Pass(\%)} \\
\midrule
R2\_text\_input\_constraints      & Missing required/maxlength/pattern/min/max on inputs       & 2,956 & 2.9 \\
R6\_dangerous\_dom\_uncertain     & Suspected unsafe DOM rendering patterns                    & 430   & 0.0 \\
R6\_dangerous\_dom\_rendering     & Dangerous DOM API usage (potential XSS)                    & 437   & 2.5 \\
R7\_continuous\_trigger\_guard    & No debounce/disable guard on repeated clicks               & 273   & 1.8 \\
R1\_sensitive\_form\_csrf         & Sensitive form lacks CSRF token                            & 191   & 0.0 \\
R1\_sensitive\_form\_post         & Sensitive form not using POST method                       & 191   & 1.0 \\
R7\_async\_error\_recovery        & Missing error recovery path for async failures             & 277   & 11.6 \\
R5\_filter\_sort\_sync            & Filter/sort state out of sync with UI or results           & 1,284 & 26.1 \\
R1\_button\_explicit\_type        & Button missing explicit type attribute                     & 1,075 & 35.5 \\
R7\_async\_loading\_state         & Missing loading indicator during async operations          & 277   & 52.0 \\
R2\_search\_trim                  & Unstable whitespace handling in search input               & 672   & 64.1 \\
R2\_invalid\_input\_handling      & Invalid input handling                                     & 1,058 & 91.8 \\
\bottomrule
\end{tabular}
\caption{
High-frequency safety check details for GPT-5.5, sorted by pass rate.
The lowest-pass checks mainly involve input constraints, unsafe DOM rendering, repeated-trigger guards, and sensitive-form protections.
}
\label{tab:app-safety-checks}
\end{table*}

\begin{table}[t]
\centering
\small
\setlength{\tabcolsep}{3pt}
\renewcommand{\arraystretch}{1.05}
\begin{tabular}{llcc}
\toprule
\textbf{Rule} & \textbf{Meaning} & \textbf{Checks} & \textbf{Pass(\%)} \\
\midrule
R7 & Async \& interaction robustness & 881   & 20.5 \\
R6 & DOM rendering safety            & 1,169 & 25.5 \\
R1 & Request security                & 1,457 & 26.4 \\
R2 & Input validation                & 5,253 & 38.8 \\
R5 & State consistency               & 2,843 & 63.8 \\
R3 & Upload security                 & 187   & 66.3 \\
R4 & Navigation security             & 45    & 97.8 \\
\bottomrule
\end{tabular}
\caption{
Safety rule-level breakdown for GPT-5.5.
The weakest rule families are asynchronous interaction robustness, DOM rendering safety, and request security.
}
\label{tab:app-safety-rules}
\end{table}

\subsection{Defect Injection Details}

We further inspect the 13 defect-injected cases where WebGen marks all test items as \textsc{Yes}, but ICG still detects the injected defect.
As shown in~\cref{tab:icg-only-error-patterns}, these cases are not dominated by visibly missing controls or rendering failures.
Instead, they involve longer-range behavioral constraints, including accumulated history preservation, cross-feature non-interference, navigation-time state retention, action gating, and pre/post state consistency.
This explains why checkpoint-style evaluation can miss them: it often verifies whether the local target appears completed, whereas ICG follows transition chains and checks requirement-linked postconditions and state invariants.
These ICG-only cases therefore show that explicit state-transition contracts provide complementary coverage for hidden state errors and cross-feature side effects beyond local checkpoint judgments.

\subsection{Safety Evaluation Details}
\label{app:safety_details}

We provide rule-level safety diagnostics for GPT-5.5, the strongest model in the main interaction evaluation.
These diagnostics are auxiliary to \bench's interaction metrics and are intended to reveal common engineering-level weaknesses in generated HTML artifacts.

As shown in \cref{tab:app-safety-rules}, the weakest rule families are asynchronous interaction robustness, DOM rendering safety, and request security.
The low pass rates for R7, R6, and R1 indicate that generated pages often miss repeated-trigger guards, safe DOM rendering practices, and basic protections for sensitive requests.
In contrast, navigation security obtains a high pass rate, but covers far fewer applicable checks and should not be interpreted as broad safety reliability.

\cref{tab:app-safety-checks} further shows that the most frequent low-pass checks involve missing input constraints, unsafe DOM rendering, repeated-click guards, and sensitive-form protections.
These results suggest that even strong MLLMs may generate functional and visually plausible webpages while omitting basic front-end safety and robustness safeguards.

\subsection{Case Study}
\label{app:case_study}

This section presents representative qualitative cases for the failure types used in our failure attribution analysis.
Each case shows the input modality, a passing artifact, a failing artifact, the executed transition, and the failed evidence.
Together, these examples show how \bench evaluates each transition from the source state to the target state and records where the expected behavior breaks.

\begin{figure*}[t]
\centering
\includegraphics[width=\textwidth]{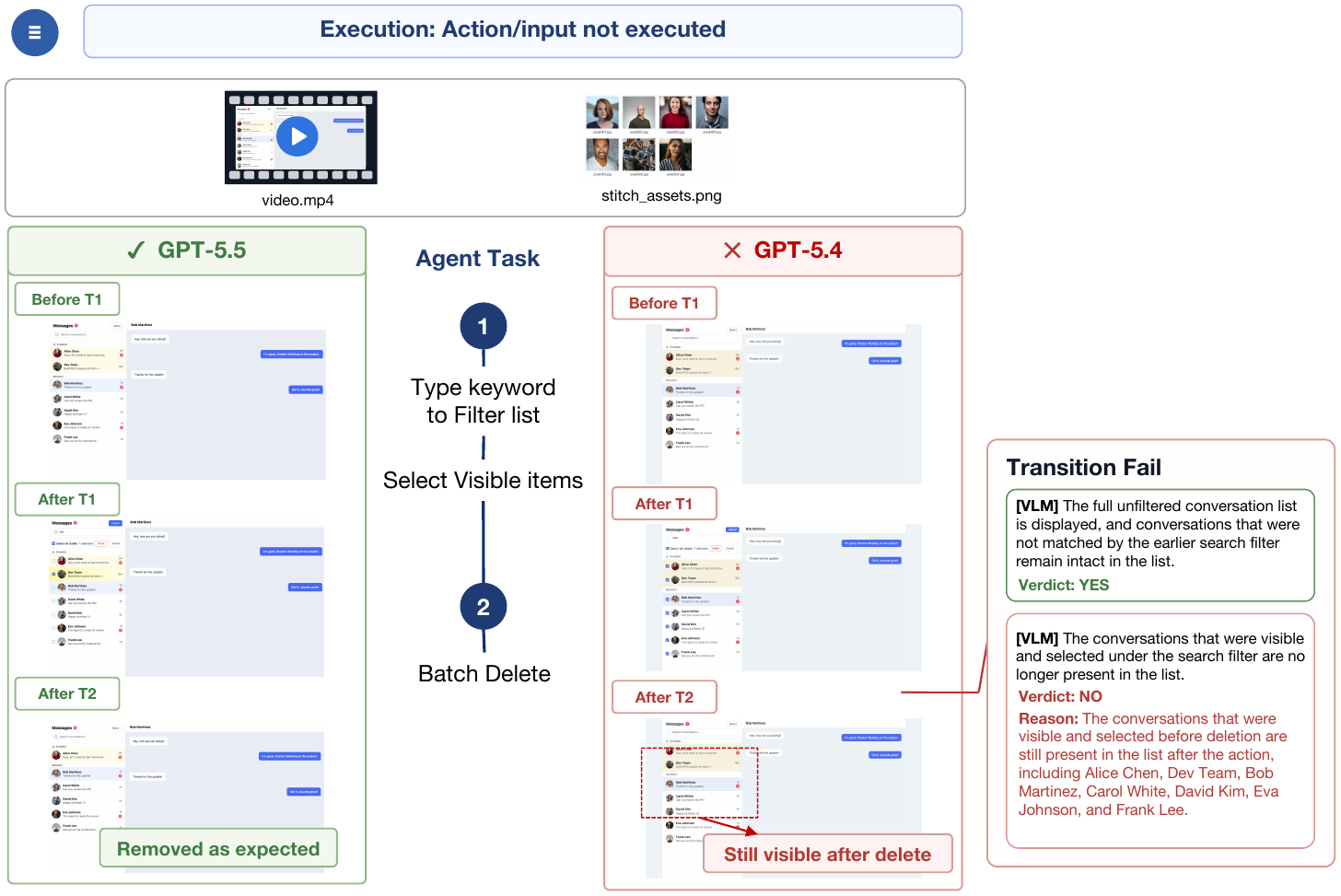}
\caption{
Execution failure in a messaging interface.
The transition requires filtering the conversation list, selecting the visible conversations, and batch deleting them.
The failing artifact keeps the selected conversations visible after deletion.
}
\label{fig:case1}
\end{figure*}

\nbf{Case 1: Execution failure.} This case tests whether a generated messaging interface can execute a batch operation after filtering and selecting visible conversations. The expected behavior is that the selected conversations disappear after the batch-delete action, while unmatched conversations remain in the restored full list. Although the failing artifact displays the search and selection flow, the selected conversations are still visible after deletion. This indicates an execution failure: the page exposes a plausible operation path, but the underlying delete action is not successfully applied to the selected items.

\begin{figure*}[t]
\centering
\includegraphics[width=\textwidth]{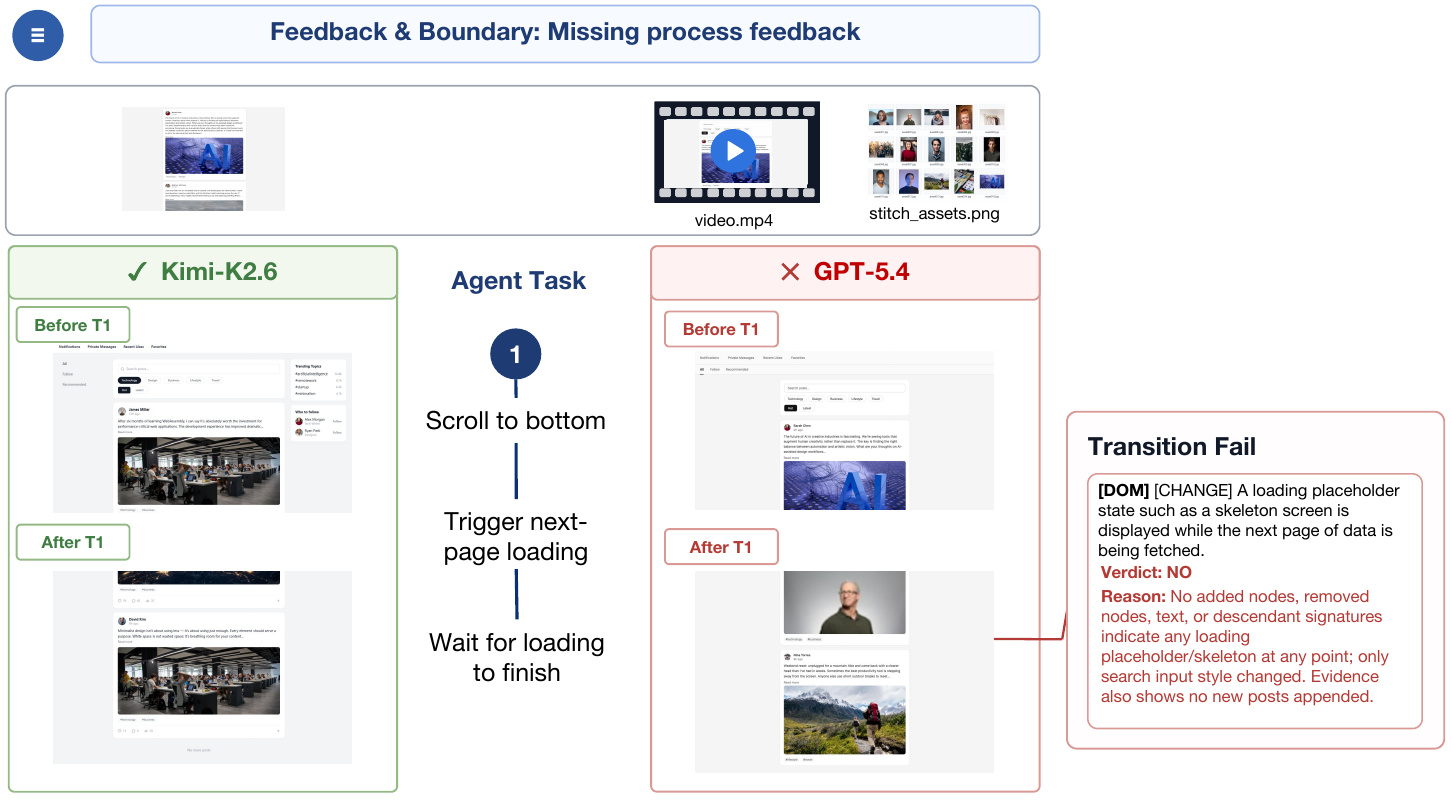}
\caption{
Feedback and boundary failure in a feed-loading interaction.
The transition requires scrolling to the bottom, triggering next-page loading, and displaying loading feedback during data fetching.
The failing artifact does not show the required loading placeholder.
}
\label{fig:case2}
\end{figure*}

\nbf{Case 2: Feedback \& Boundary failure.}
This case focuses on process feedback during an infinite-scroll interaction.
After the user scrolls to the bottom, the page should indicate that the next page of content is being fetched, for example through a skeleton screen or loading placeholder.
The failing artifact reaches the scroll boundary but provides no observable loading state, and the evidence also shows no newly appended posts.
This failure shows that the main interaction entry point may exist, while the boundary-state feedback required for a realistic web interaction is still missing.

\begin{figure*}[t]
\centering
\includegraphics[width=\textwidth]{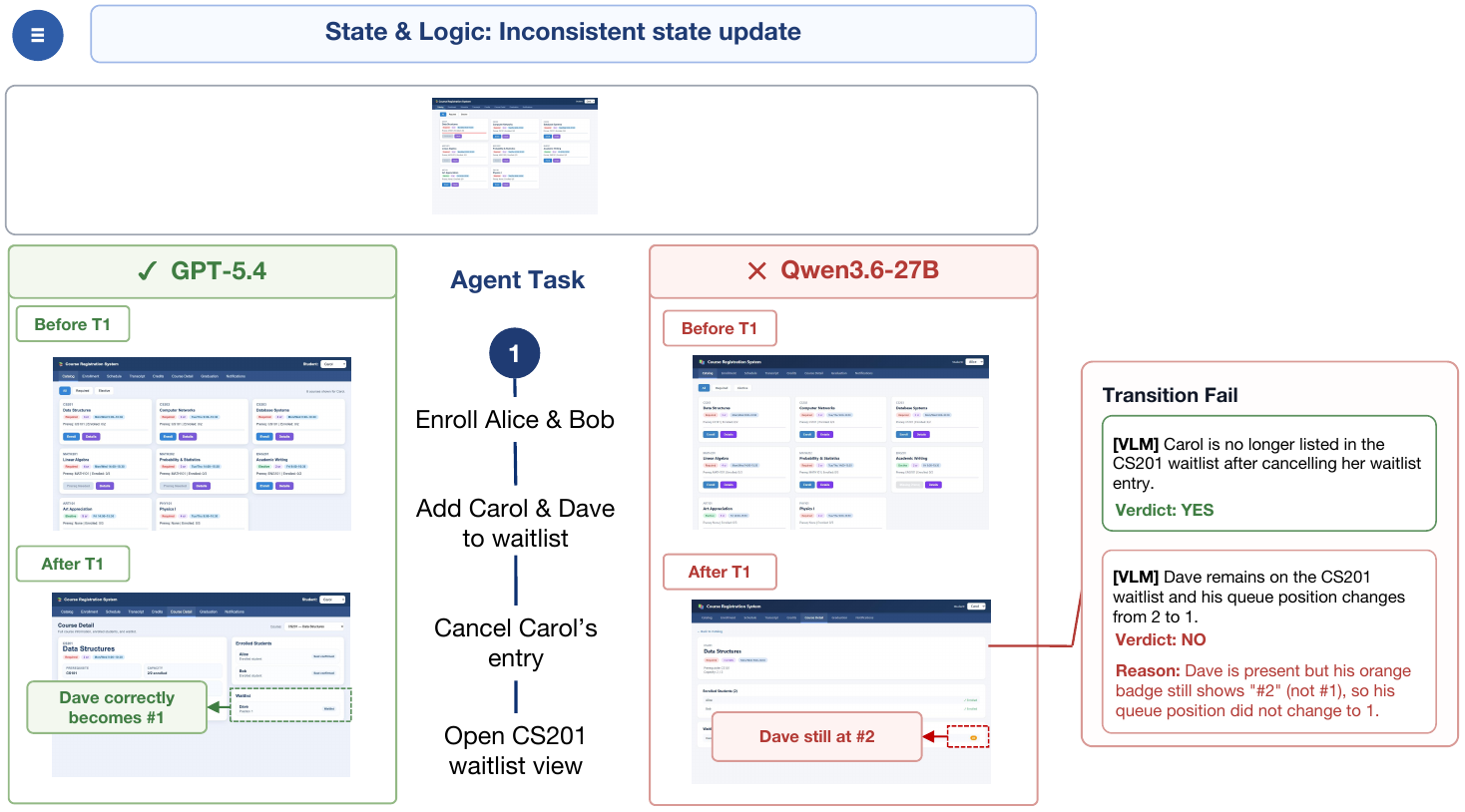}
\caption{
State-and-logic failure in a course waitlist interaction.
After Carol's waitlist entry is cancelled, Dave should remain on the CS201 waitlist and move from position \#2 to \#1.
The failing artifact removes Carol but leaves Dave's queue position unchanged.
}
\label{fig:case3}
\end{figure*}

\nbf{Case 3: State \& Logic failure -- inconsistent state update.}
This case evaluates whether a course registration page correctly updates dependent waitlist state.
The transition first enrolls Alice and Bob, adds Carol and Dave to the CS201 waitlist, cancels Carol's entry, and then opens the waitlist view.
The failing artifact correctly removes Carol, but Dave remains marked as \#2 instead of being promoted to \#1.
The error is an incomplete state update: one part of the state changes, while the dependent queue order is left stale.

\begin{figure*}[t]
\centering
\includegraphics[width=\textwidth]{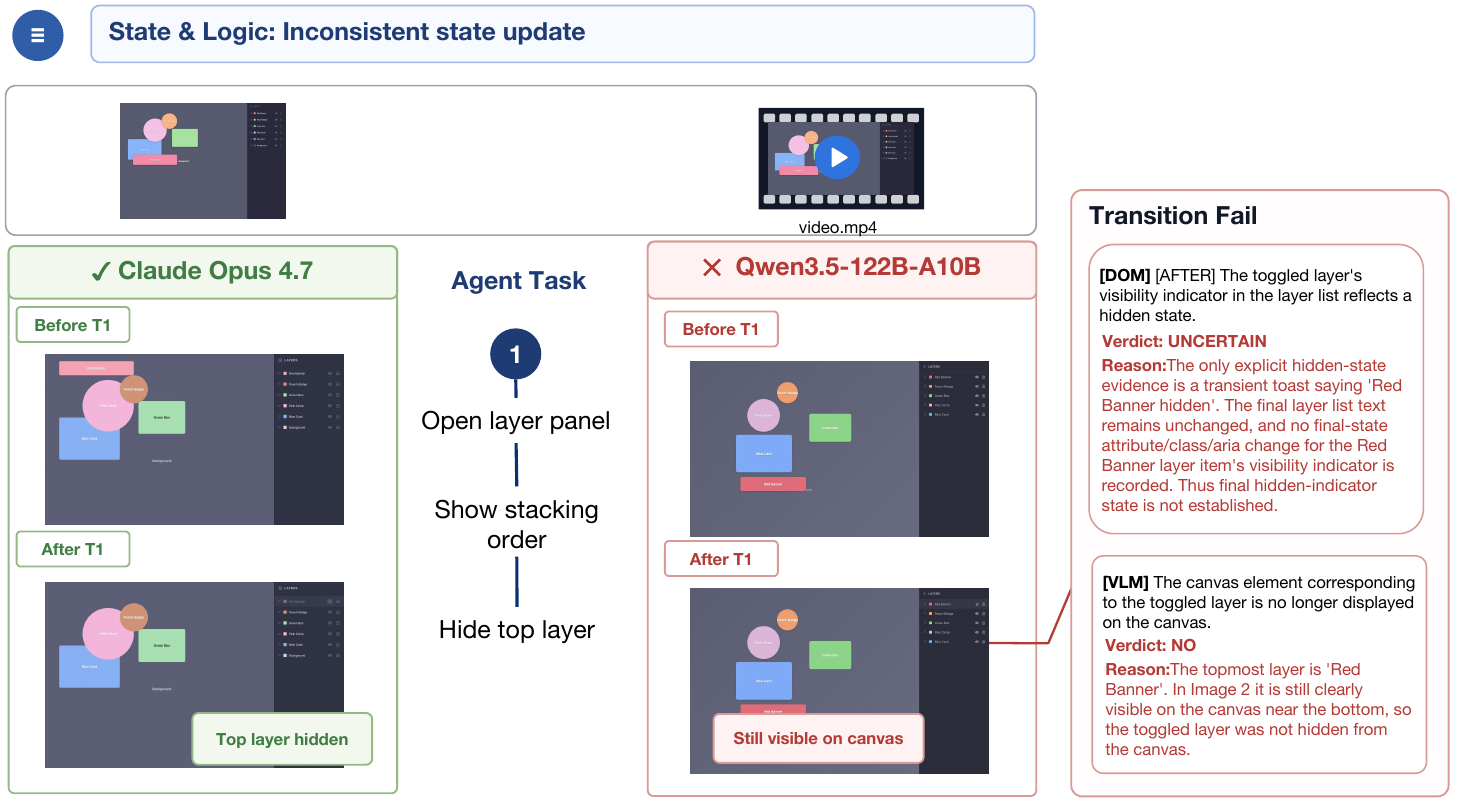}
\caption{
State-and-logic failure in a layer-list interaction.
The transition requires opening the layer panel and hiding the topmost layer.
The failing artifact gives weak or transient hidden-state evidence but leaves the corresponding canvas element visible.
}
\label{fig:case4}
\end{figure*}

\nbf{Case 4: State \& Logic failure -- cross-view inconsistency.}
This case tests synchronization between a layer list and the visible canvas.
After the topmost layer is hidden from the layer panel, the corresponding object should no longer appear on the canvas, and the layer list should reflect the hidden state.
The failing artifact provides only uncertain final-state evidence in the layer list and still displays the hidden layer on the canvas.
This exposes a cross-view state inconsistency: the control-side state and the rendered canvas state are not synchronized.

\begin{figure*}[t]
\centering
\includegraphics[width=\textwidth]{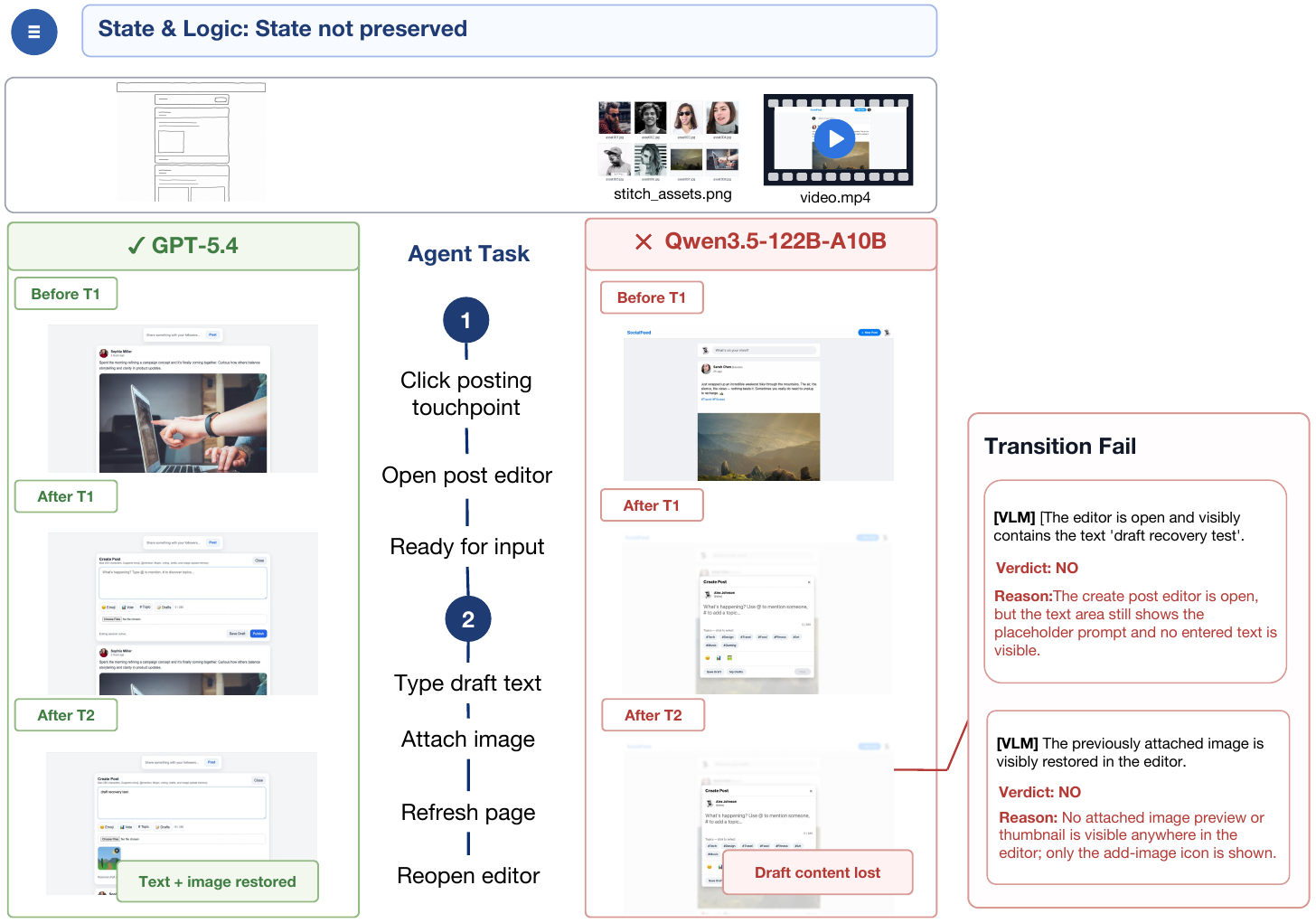}
\caption{
State preservation failure in a draft-recovery workflow.
After typing text, attaching an image, refreshing the page, and reopening the editor, the draft should be restored.
The failing artifact loses both the entered text and the attached image.
}
\label{fig:case5}
\end{figure*}

\nbf{Case 5: State \& Logic failure -- state not preserved.}
This case examines whether a social post editor preserves draft content across an unexpected refresh.
The transition requires opening the editor, entering the text ``draft recovery test'', attaching an image, refreshing the page, and reopening the editor.
The failing artifact reopens the editor but shows the placeholder text and no image preview, meaning that neither the text nor the attachment is restored.
This demonstrates a persistence failure: the interaction is locally available, but the generated page does not preserve user-created state across the page lifecycle.

\begin{figure*}[t]
\centering
\includegraphics[width=\textwidth]{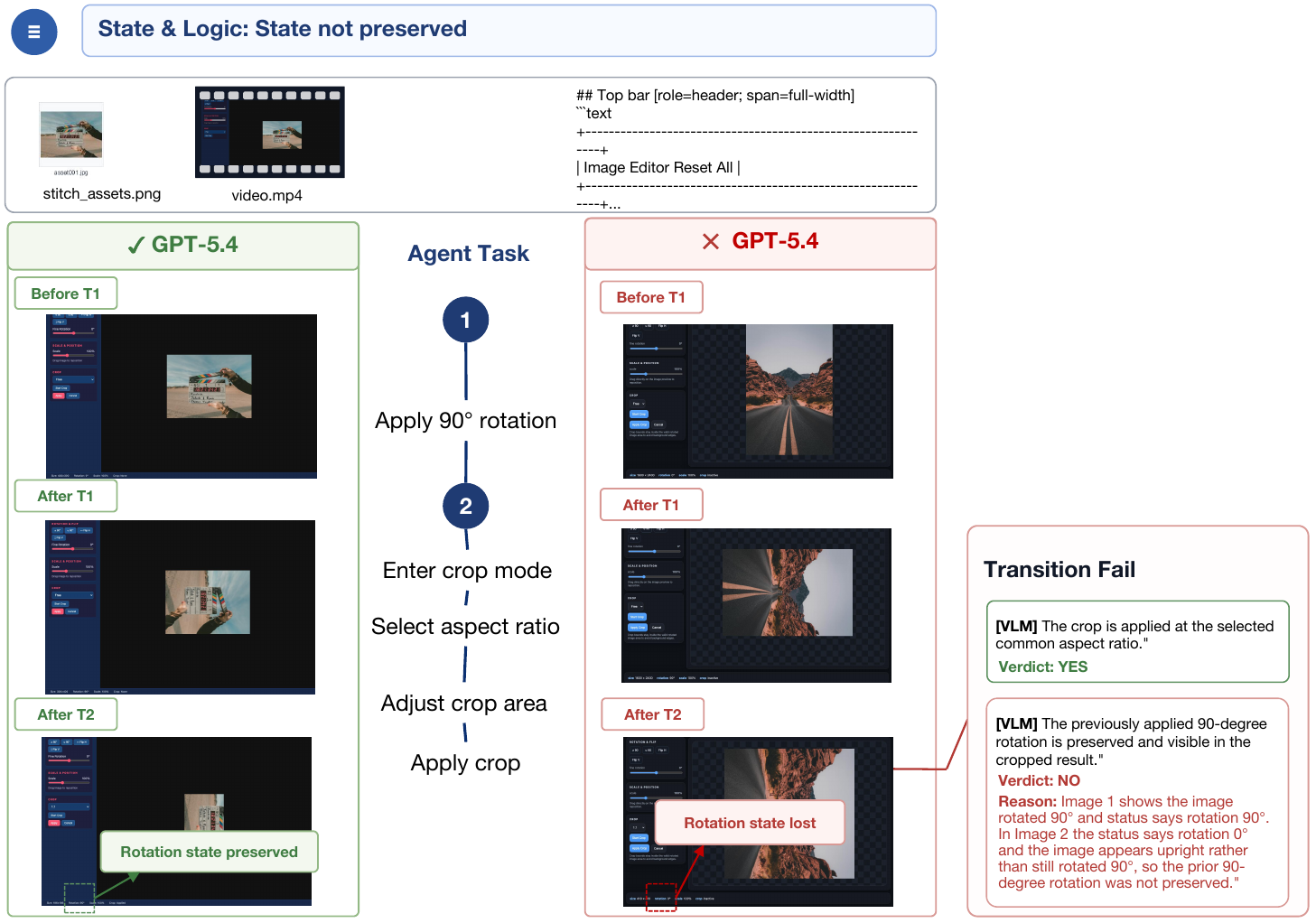}
\caption{
State preservation failure in an image-editing workflow.
The transition requires applying a 90-degree rotation, entering crop mode, selecting an aspect ratio, and applying the crop while preserving the prior rotation.
The failing artifact applies the crop but resets the rotation state.
}
\label{fig:case6}
\end{figure*}

\nbf{Case 6: State \& Logic failure -- operation state reset.}
This case evaluates whether an image editor preserves earlier editing state when a later operation is applied.
The transition first rotates the image by 90 degrees and then performs a crop with a selected aspect ratio.
The failing artifact applies the crop, but the final result no longer preserves the prior 90-degree rotation state.
This is a state-preservation error across sequential editing operations: the later crop operation incorrectly resets an earlier transformation state.

\section{Prompt Templates}
\label{app:prompt_templates}
This section lists the prompt templates used in \bench, including templates for test data contract generation, test item generation, Interaction Contract Graph construction, contract-guided agent execution, DOM assertion scoring, visual postcondition scoring.

\begin{PromptBox}
{Test Data Contract Generation}
{Prompt for deriving initial functional readiness from requirements.}
{fig:prompt-test-data-contract}

You are a frontend test specification designer.
Given a requirement list, produce a minimal test data contract describing what the page must be functionally ready to do on first load.
No reference implementation is provided; derive the contract from the requirements alone.

\textbf{Rules:}
\begin{enumerate}[nosep, leftmargin=*, itemsep=2pt]
\item Describe functional readiness, not UI structure, visual layout, DOM hierarchy, or exact element counts.
\item For multi-page, multi-view, tabbed, wizard, or navigation-based apps, explicitly name the initial page, view, route, or mode shown on first load.
\item Do not prescribe positions, component hierarchy, styling, specific mock values, asset sources, or exact numbers of items.
\item Include only conditions needed for the first test action to be possible.
\item Do not expose implicit requirements. Remove contract text that reveals behavior beyond the explicit requirements.
\item If a default initial view is not specified but the app requires one, choose a reasonable primary workflow view and state it functionally.
\end{enumerate}

\textbf{Output:} Return exactly one JSON object:
{\small\texttt{\{"test\_data\_contract": "functional preconditions describing page readiness"\}}}
\end{PromptBox}

\begin{PromptBox}
{Test Item Generation}
{Prompt for converting requirements into implementation-neutral test items.}
{fig:prompt-test-item-generation}

You are a frontend test specification designer.
Produce a test item list covering every testable behavior in the given requirement list.
No reference implementation is provided; derive the test items from the requirements alone.

\textbf{Rules:}
\begin{enumerate}[nosep, leftmargin=*, itemsep=2pt]
\item Generate one test item per distinct testable behavior. Every explicit and implicit requirement ID must appear in at least one item's \texttt{req\_ids}.
\item Triggers and expected results must be implementation-neutral. A trigger describes user intent, not a click sequence or pure observation; an expected result describes the semantic outcome.
\item Combine tightly coupled behaviors that share the same trigger and artifact, but split named cases with distinct outcomes.
\item Use only primary requirement IDs in \texttt{req\_ids}, normally at most two per item. If an expected result verifies a requirement, include that requirement ID.
\item Do not invent behaviors. Failure, error, boundary, and follow-up cases must be grounded in named requirements.
\item Implicit requirements may refine explicitly stated behaviors, but must not introduce new scenarios by themselves.
\item Do not create standalone negative-capability items; fold unavailable actions into the expected result of the state-changing item that causes them.
\item Selection-gated controls are modeled as one behavior: the trigger selects the required content and invokes the control, while the expected result covers the gated availability.
\item Guard or prevention behavior may be a separate item only when it has a distinct user trigger and a clearly observable prevention outcome.
\item For toggles or bidirectional behavior, describe switching from the current state to the alternative without assuming a default.
\item Do not split countdown, cooldown, or expiry flows across separate items; keep the complete timed user intent in one item.
\end{enumerate}

\textbf{Output:} Return exactly one JSON object:
{\small\texttt{\{"test\_items": [\{"item\_id": "TI-1", "req\_ids": [...], "description": "...", "trigger": "...", "expected\_result": "..."\}]\}}}
\end{PromptBox}

\begin{PromptBox}
{ICG Generation}
{Prompt for generating the state-transition Interaction Contract Graph.}
{fig:prompt-icg-generation}

You are a frontend interaction test case designer.
Generate a test specification with \texttt{states} and \texttt{transitions}.
Each transition specifies a source state, a target state, a self-contained \texttt{agent\_task}, mapped test item IDs, \texttt{dom\_assertions}, and/or visual \texttt{postconditions}.

\textbf{State rules:}
\begin{enumerate}[nosep, leftmargin=*, itemsep=2pt]
\item States are stable, replayable page checkpoints. Use \texttt{S0}, \texttt{S1}, etc.; \texttt{S0} is the initial state.
\item State descriptions are short, implementation-neutral page snapshots. Do not model transient UI such as spinners, toasts, timers, or animations as states.
\item Define a new state whenever visible content, selected controls, open panels, input values, displayed artifacts, or persistent UI state differs materially.
\item Preserve state continuity: unchanged visible aspects from the source state should carry into the target state's description.
\item A self-loop is valid only when the after-state is observably identical to the before-state.
\end{enumerate}

\textbf{Transition rules:}
\begin{enumerate}[nosep, leftmargin=*, itemsep=2pt]
\item Use sequential IDs \texttt{T1}, \texttt{T2}, etc. Each transition declares \texttt{from}, \texttt{to}, \texttt{agent\_task}, \texttt{mapped\_test\_items}, and at least one non-empty assertion list.
\item \texttt{preconditions} are allowed only on \texttt{T1}; later transitions must not contain preconditions.
\item Use \texttt{dom\_assertions} for DOM mutation evidence, temporal evidence, and element-level state. Prefix each DOM assertion with exactly \texttt{[CHANGE]} or \texttt{[AFTER]}.
\item Use visual \texttt{postconditions} for outcomes judged from before/after screenshots. Do not prefix postconditions.
\item The \texttt{agent\_task} must describe the user's goal from the current \texttt{from} state. It must be actionable, self-contained, and not a low-level selector or click sequence.
\item The \texttt{agent\_task} must not refer to previous transitions; if prior context is needed, describe it as a persistent property of the current source state.
\item Do not create observation-only or self-check transitions. Static checks should be attached to the transition that establishes the checked state.
\item No hitchhiking: every mapped test item must be directly caused by this transition's \texttt{agent\_task} and verifiable from this transition's final state.
\item Every input test item must be covered by at least one transition.
\item Every mapped test item must have at least one direct \texttt{dom\_assertion} or visual \texttt{postcondition} in the same transition.
\item Independent features should fan out from the same source state rather than being falsely chained; serial chains are used only when the later transition genuinely requires the prior target state.
\item Compound transitions may combine two or three operations only when they share the same artifact and are intended to test state interference or clobbering.
\item Do not invent UI controls, defaults, failure paths, labels, data values, or data assumptions not supported by the Test Data Contract or test items.
\item Empty-state tests must explicitly clear or remove pre-existing content when the Test Data Contract does not guarantee emptiness.
\end{enumerate}

\textbf{Output:} Return exactly one JSON object with top-level fields \texttt{states} and \texttt{transitions}. Do not include markdown fences or extra prose.
\end{PromptBox}

\begin{PromptBox}
{Agent Execution}
{Prompt used by the browser agent to execute one transition.}
{fig:prompt-agent-execution}

You are a robot browsing the web to execute a web-testing task.
In each iteration, you receive an indexed DOM observation where elements are prefixed with \texttt{[N]}, and newly appeared elements may be marked with \texttt{*[N]}.
Choose exactly one action using indices from the latest observation.

\textbf{Action grammar:}
\begin{quote}
\small
\texttt{Click [N]} \\
\texttt{Click [N]; count} \\
\texttt{Dismiss} \\
\texttt{DoubleClick [N]} \\
\texttt{RightClick [N]} \\
\texttt{LongPress [N]; ms} \\
\texttt{Hover [N]} \\
\texttt{Input [N]; text} \\
\texttt{InputDate [N]; YYYY-MM-DD} \\
\texttt{Clear [N]} \\
\texttt{Blur [N]} \\
\texttt{Select [N]; option label} \\
\texttt{Check [N]} / \texttt{Uncheck [N]} \\
\texttt{Press [N]; key} \\
\texttt{Press [N]; key; count} \\
\texttt{Scroll [N or WINDOW]; up|down|top|bottom|left|right} \\
\texttt{Drag [N]; [M]} \\
\texttt{Drag [N]; offset\_x=px,offset\_y=px} \\
\texttt{DragRange [N]; target value} \\
\texttt{ClickAt; x=px y=px} \\
\texttt{Upload [N]; file\_path} \\
\texttt{Upload [N]; file\_path|file\_path} \\
\texttt{SelectText [N]; text to select} \\
\texttt{Wait; ms} \\
\texttt{Refresh} \\
\texttt{GoBack} \\
\texttt{Reset} \\
\texttt{Done}
\end{quote}

\textbf{Execution constraints:}
Use only provided test assets for upload, do not bypass a required interaction path with a similar final state, and emit \texttt{Done} only after the requested user action has fully executed.

\textbf{Output:} Return exactly one JSON object:
{\small\texttt{\{"thought": "brief reasoning", "action": "ONE Action"\}}}
\end{PromptBox}

\begin{PromptBox}
{DOM Assertion Scoring}
{Prompt for judging DOM assertions from mutation evidence.}
{fig:prompt-dom-scoring}

You are a strict UI test evaluator.
Determine whether DOM assertions are satisfied based on structured DOM event evidence collected during a web interaction.
The evidence contains the action performed, initial and final DOM snapshots, mutation events, changed attributes, added and removed nodes, and interactive-element summaries.

\textbf{Assertion semantics:}
\begin{enumerate}[nosep, leftmargin=*, itemsep=2pt]
\item \texttt{[CHANGE]} means the condition appeared at any point in the full timeline, including initial snapshot, mutation events, intermediate summaries, or final snapshot.
\item \texttt{[AFTER]} means the condition must hold in the final stable state. Timeline evidence may help locate the target, but the final state must satisfy the assertion.
\end{enumerate}

\textbf{Judging rules:}
\begin{enumerate}[nosep, leftmargin=*, itemsep=2pt]
\item Locate elements by semantic role, tag, ID, class, text, attribute changes, and interactive-element summaries.
\item Hidden or \texttt{not-visible} text counts as DOM evidence but not as proof that the text is visibly present.
\item For disabled or non-interactive state, prioritize \texttt{pointer-events: none}, native \texttt{disabled} or \texttt{aria-disabled}, then state-indicative class tokens.
\item For selected, active, highlighted, expanded, checked, or pressed states, use ARIA fields first and class tokens as secondary evidence.
\item Added and removed nodes may prove transient feedback such as loading, saving, progress, confirmation, or disappearance.
\item Judge by semantic equivalence rather than exact wording, but be strict on factual contradictions.
\item Prefer \texttt{UNCERTAIN} when evidence is incomplete, except when absence from final interactive elements directly supports a non-interactive or absent assertion.
\item For debounce or delayed-update assertions, accept evidence of a single delayed update after input settles rather than requiring keystroke-level events.
\item Do not treat hidden template text as evidence that a visible status or control is active.
\end{enumerate}

\textbf{Output:} Return exactly one JSON object:
{\small\texttt{\{"evaluations": [\{"think": "...", "result": "YES|NO|UNCERTAIN"\}]\}}}
\end{PromptBox}

\begin{PromptBox}
{Visual Postcondition Scoring}
{Prompt for judging postconditions from before/after screenshots.}
{fig:prompt-visual-postcondition-scoring}

You are a strict UI test evaluator.
Compare two screenshots: Image 1 is before the interaction, and Image 2 is after the interaction.
Determine whether each assertion holds on the current page.

\textbf{Judging rules:}
\begin{enumerate}[nosep, leftmargin=*, itemsep=2pt]
\item Judge by semantic equivalence rather than exact wording; be strict on factual correctness but lenient on terminology.
\item For conditional assertions, determine which branch applies from the screenshots and evaluate only that branch.
\item If content is clipped by the viewport or a scrollable container, evaluate only fully visible items.
\item Accept small numeric changes hidden by rounding or abbreviation when the structural outcome is otherwise correct.
\item For search or filter assertions, an empty result may pass if the filter is visibly active and the page shows a valid empty state.
\item For body or full-text search, do not require every visible result row to display the matching keyword if the active query and changed result set support the outcome.
\item For visually ambiguous natural-image flips, do not answer \texttt{UNCERTAIN} solely because the flip is hard to distinguish when other requested edits are clearly visible.
\item Use before/after differences to judge reordering, expansion, collapsed panels, drag placement, list updates, and stale-state removal.
\end{enumerate}

\textbf{Output:} Return exactly one JSON object:
{\small\texttt{\{"evaluations": [\{"think": "...", "result": "YES|NO|UNCERTAIN"\}]\}}}
\end{PromptBox}

\section{Code and Data Availability}
\label{sec:availability}

Upon acceptance, we will release the code and data for \bench under the MIT license.
The release will include task specifications, requirement annotations, Interaction Contract Graphs, evaluation scripts, prompt templates, and aggregated results for reproducing the main experiments.
We will exclude information that may identify individual contributors or annotators for privacy reasons.

\end{document}